\documentclass{article}

\usepackage[preprint]{neurips_2026}
\usepackage[utf8]{inputenc}
\usepackage[T1]{fontenc}
\usepackage{url}
\usepackage{booktabs}
\usepackage{amsfonts}
\usepackage{nicefrac}
\usepackage{microtype}
\usepackage{xcolor}
\usepackage{hyperref}

\usepackage{graphicx}
\usepackage{bm}
\usepackage{wrapfig}
\usepackage{tcolorbox}
\tcbuselibrary{breakable}
\usepackage{amsmath,amssymb}
\usepackage{subcaption}
\usepackage{placeins}
\usepackage{float}
\usepackage{amsthm}
\usepackage{titletoc}
\newtheorem{theorem}{Theorem}
\newtheorem{proposition}{Proposition}

\usepackage[table]{xcolor}
\usepackage{colortbl}
\usepackage{placeins}
\tcbuselibrary{listings,breakable,skins}

\definecolor{deltagray}{RGB}{242,242,242}
\definecolor{focalgreen}{RGB}{214,232,208}

\newcommand{\focalrowbg}{\rlap{\smash{\color{focalgreen!50}\rule[-0.30em]{0.90\textwidth}{1.35em}}}}
\newcommand{\bcell}[1]{\textcolor{gray}{#1}}

\usepackage{adjustbox}

\newcommand{\expfigvspace}{\vspace{-0.8em}}
\newcommand{\expcaptionsetup}{\captionsetup{font=small,skip=2pt}}

\usepackage{enumitem}
\usepackage{multirow}
\usepackage[table,dvipsnames]{xcolor}

\title{Focal Reward: Balanced Reinforcement Learning under Rubric-Based Rewards}

\author{
\textbf{Yu Huang}\textsuperscript{1,*} \quad
\textbf{Zihua Zhao}\textsuperscript{1,*} \quad
\textbf{Zhaoxin Huan}\textsuperscript{2} \quad
\textbf{Wanli Gu}\textsuperscript{2} \quad
\textbf{Feng Hong}\textsuperscript{1} \quad
\textbf{Xinmu Ge}\textsuperscript{1}
\\
\textbf{Lin Yuan}\textsuperscript{2} \quad
\textbf{Weichang Wu}\textsuperscript{2} \quad
\textbf{Qiang Hu}\textsuperscript{1} \quad
\textbf{Xiaolu Zhang}\textsuperscript{2} \quad
\textbf{Jun Zhou}\textsuperscript{2,\textdagger} \quad
\textbf{Jiangchao Yao}\textsuperscript{1,\textdagger}
\\
\textsuperscript{1}Shanghai Jiao Tong University
\quad
\textsuperscript{2}Ant Group
\\
\textsuperscript{*}Equal contribution.
\quad
\textsuperscript{\textdagger}Corresponding authors.
}

\begin{document}

\maketitle

\begin{abstract}
The open-ended generation in LLMs usually requires multi-dimensional rubrics to adequately assess quality and guide the improvement of reinforcement learning. However, a critical dilemma inherent in this training paradigm is the imbalanced reward polarization along different rubric dimensions. Under this bottleneck, even if LLMs achieve relatively high rewards after training, they may still exhibit severe deficiencies in certain dimensions, leading to a direct deterioration in user experience. To address this problem, we propose \emph{Focal Reward}, a novel objective to automatically balance the training of reinforcement learning under rubric-based rewards. Specifically, we first leverage an inverse reward projection mechanism to estimate the saturation degree of each criterion in the rubric, which forms the basis to calibrate the reward direction. Then, the final objective is designed with an automatically reweighting coefficient for each criterion to achieve the fine-grained balancing. Extensive experiments across three model scales and six benchmarks demonstrate that our \emph{Focal Reward} method outperforms the strongest static aggregation baseline in all 18 model--benchmark comparisons. Rollout, mechanism, and ablation analyses further show that these gains arise from online, saturation-aware reallocation toward rubrics that still have room for improvement.
\end{abstract}



\section{Introduction}
\label{sec:intro}

Unlike domains such as mathematics or coding~\citep{shao2024deepseekmath,liu2025code}, the open-ended generation capabilities of LLMs are often subjective and inherently multi-dimensional, making them difficult to evaluate reliably~\citep{su2025crossing,jia2025writing}. As a result, modern alignment paradigms typically explore rubric-based AI judges that provide structured, criterion-level feedback, rather than depending solely on holistic or opaque preference signals~\citep{DBLP:conf/icml/0001PMMFLBHCRP24,DBLP:journals/corr/abs-2507-17746,DBLP:journals/corr/abs-2507-18624}. The resulting advances provide a natural interface for optimizing LLMs with the diverse qualities expected in open-ended generation~\citep{DBLP:journals/corr/abs-2508-12790}.


For rubric-based reinforcement learning, two issues are central: 1) constructing appropriate rubrics to support fine-grained judgment; 2) aggregating criterion-level feedback for policy optimization. For the former, prior work has explored personalized rubrics, training-aligned rubric design, and group-relative evaluation~\citep{DBLP:journals/corr/abs-2510-07743,DBLP:conf/acl/GuptaSZMGBYR25,DBLP:journals/corr/abs-2602-14069,DBLP:conf/iclr/KimS0JLLYSKTS24,DBLP:conf/iclr/ZhuWW25}. For the latter, most existing methods use static weighted aggregation to merge criterion-level feedback into a scalar reward~\citep{DBLP:journals/corr/abs-2507-17746,DBLP:journals/corr/abs-2507-18624,DBLP:journals/corr/abs-2510-07743}. Despite their effectiveness, static scalarizers can lead to imbalanced optimization across rubric criteria, with some criteria becoming saturated while others remain under-satisfied. This imbalance can produce high-reward but behaviorally flawed generations. In such cases, surface-level quality may receive more consistent reward than decisive task-specific constraints, causing models to over-optimize easy-to-score aspects while failing to satisfy the user's actual request.

\begin{figure*}[t]
    \centering
    \expcaptionsetup
    \includegraphics[width=\linewidth]{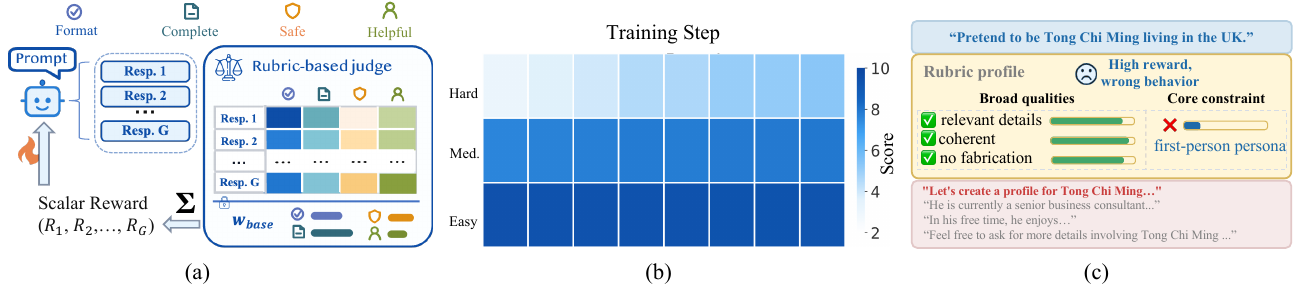}
    \caption{
    Motivation and intuition of \emph{Focal Reward}:
    (a) Rubric-based RL evaluates rollouts across criteria and aggregates criterion scores into scalar rewards;
    (b) criteria at different difficulty levels exhibit distinct saturation trajectories, as easier criteria saturate earlier while harder ones remain under-satisfied;
    (c) static aggregation can favor high-reward yet flawed rollouts, where broad quality is satisfied but a decisive requirement is missed.
    }
    \label{fig:intro_overview}
    \expfigvspace
\end{figure*}

As shown in Figure~\ref{fig:intro_overview}, rubric-based RL optimizes LLMs through diverse criteria spanning different difficulty levels, and they achieve different saturation degrees throughout the training. Easier criteria, such as fluency or format compliance, may become satisfied early, whereas harder criteria, such as subtle trade-offs or nuanced helpfulness, often remain under-satisfied for longer. Static scalarizers can fail to track this changing profile, continuing to emphasize saturated criteria while under-weighting those with substantial remaining headroom. This can rank a polished rollout higher when broad-quality gains dominate the scalar reward margin, even if a decisive task-specific requirement is unmet. Thus, the key issue lies in how criterion-level feedback is scalarized before policy optimization.

To address this failure mode of static scalarization, we propose \emph{Focal Reward}, a novel objective for automatically balancing reinforcement learning under rubric-based rewards. Focal Reward estimates criterion saturation through inverse reward projection and uses it to calibrate the reward direction: saturated criteria receive reduced emphasis, whereas high-headroom criteria receive stronger training signals. It then assigns adaptive reweighting coefficients before synthesizing criterion-level feedback into scalar reward for policy optimization. Importantly, Focal Reward leaves the judge, rubric, and policy optimization framework unchanged, restricting the intervention to reward synthesis.

In summary, our main contributions are as follows:

\begin{itemize}[leftmargin=1.5em]

\item We identify \emph{imbalanced reward polarization} in rubric-based reinforcement learning. 
When rubric criteria saturate at different rates, static scalarizers can continue rewarding already-satisfied dimensions while under-emphasizing criteria that still retain substantial room for improvement.

\item  We propose \emph{Focal Reward}, a novel objective for automatically balancing reinforcement learning under rubric-based rewards, with theoretical grounding. It uses inverse reward projection to estimate criterion saturation, then adaptively reweights criteria by reducing emphasis on saturated criteria and increasing emphasis on criteria with larger remaining headroom.

\item We validate Focal Reward across three model scales and six benchmarks, where it outperforms the strongest static aggregation baseline in all 18 model--benchmark comparisons. Average gains over the strongest static baseline range from +1.38 to +2.50 points across model scales, with rollout, mechanism, and ablation analyses supporting online saturation-aware reallocation.

\end{itemize}



\section{Preliminaries}
\label{sec:preliminaries}

A typical rubric-based reinforcement learning for open-ended generation consists of two central parts, rubric-based judgment and reward aggregation for policy optimization, detailed as follows.





\noindent
\textbf{Rubric-based Judgment.} 
Given an LLM policy \(\pi_\theta\) parameterized by \(\theta\) and a dataset \(\mathcal{D}\) of open-ended prompts, the policy autoregressively generates a group of \(G\) candidate rollouts \(Y=\{y_i\}_{i=1}^G \sim \pi_\theta(\cdot \mid x)\) for each prompt \(x \in \mathcal{D}\). Unlike verifiable tasks, open-ended generation is implicitly evaluated in multiple dimensions. Specifically, the typical rubric-based judgment introduces a rubric of \(K\) criteria \(\mathcal{C}=\{c^{(k)}\}_{k=1}^{K}\), and an LLM judge provides criterion-level feedback for each \(y_i\). For an ordered comparison between two sampled rollouts \(y_i\) and \(y_j\), where \(y_j\) serves as the reference, the judge scores \(y_i\) and returns a criterion-level score vector
$\bm{s}_{i,j}=J(x,y_i,y_j;\mathcal{C}) \in [0,S_{\max}]^K.$
Its \(k\)-th entry \(s_{i,j}^{(k)}\) denotes the score assigned to \(y_i\) on criterion \(c^{(k)}\), and \(S_{\max}\) is usually predefined, \textit{e.g.,} $10$.

\noindent
\textbf{Reward Aggregation.} To turn criterion-level feedback into a scalar training signal, existing methods mainly rely on a fixed scalarizer, implemented as \textit{pre-defined or uniform weights} over criteria. Let $\bm w_{\mathrm{base}}=\bigl(w_{\mathrm{base}}^{(1)},\dots,w_{\mathrm{base}}^{(K)}\bigr)$ denote the static weight vector, and then the judgment is computed as
\begin{equation}
S_{i,j}(\bm{w})=\langle \bm{w},\bm{s}_{i,j}\rangle=\sum_{k=1}^{K} w^{(k)} s_{i,j}^{(k)},\qquad
\Delta_{i,j}(\bm{w})=S_{i,j}(\bm{w})-S_{j,i}(\bm{w}). \label{eq:delta}
\end{equation}
The resulting reward is then constructed by aggregating the scores credited to the corresponding pairwise comparisons within the sampled group~\citep{DBLP:journals/corr/abs-2601-06487}, which is formulated as
\begin{equation}
R_i(\bm w)=\sum_{j\ne i}\Phi_\tau(\Delta_{i,j}(\bm w)),\qquad
\Phi_\tau(\Delta)=
\begin{cases}
2\,\mathrm{sgn}(\Delta), & |\Delta|\ge\tau,\\
\mathrm{sgn}(\Delta), & 0<|\Delta|<\tau,\\
0, & \Delta=0 .
\end{cases}
\label{eq:group_reward}
\end{equation}
where $\mathrm{sgn}(\cdot)$ is the sign function and $\tau > 0$ denotes the strong-preference threshold. $\Phi_\tau(\Delta)$ acts as a scoring function, converting each pairwise margin $\Delta_{i,j}(\bm{w})$
into a discrete win/loss outcome. Following GRPO~\citep{shao2024deepseekmath}, the scalar reward $R_i$ of each rollout is converted into a group-relative advantage $\hat A_i$, and updates the policy using a following regularized clipped objective
{\small
\begin{equation}
\label{eq:grpo_obj}
\begin{aligned}
\mathcal{J}_{\mathrm{GRPO}}(\theta)
&=
\mathbb{E}_{x\sim \mathcal{D},\{y_i\}_{i=1}^{G}\sim\pi_{\theta_{\mathrm{old}}}(\cdot\mid x)}
\\
&\quad\left[
\frac{1}{G}\sum_{i=1}^{G}\frac{1}{|y_i|}
\sum_{t=1}^{|y_i|}
\left(
\min\!\big(r_{i,t}(\theta)\hat{A}_i,\,
\mathrm{clip}(r_{i,t}(\theta),1-\epsilon,1+\epsilon)\hat{A}_i\big)
-\beta D_{\mathrm{KL}}(\pi_\theta\|\pi_{\mathrm{ref}})
\right)
\right].
\end{aligned}
\end{equation}
}
Here, $r_{i,t}(\theta)$ is the token-level importance ratio against the pre-update policy and $D_{\mathrm{KL}}$ computes the KL penalty. Additional details of the policy optimization are provided in Appendix~\ref{app:details_GRPO}. Under this framework, policy updates are ultimately driven by the scalar rewards assigned to sampled rollouts.

\section{Method}
\label{sec:method}

As analyzed in the introduction, directly optimizing the reward acquired by static scalarization will induce the imbalanced reward polarization among different rubric criteria. To address this challenge, we introduce a novel \emph{Focal Reward} method to dynamically calibrate the emphasis in reinforcement learning, which we detail in this part, followed by a theoretical analysis.

\subsection{Focal Reward}
\label{sec:focal_reward}

Our idea is inspired by focal loss~\citep{DBLP:conf/iccv/LinGGHD17}, where class-wise weights are dynamically adjusted to emphasize overlooked categories. But in technical design, our case is not as straightforward as the focal loss. For clarity, we first use inverse reward projection to estimate the saturation degree of each criterion in the rubric, and then use the saturation degree to dynamically rebalance the static weight, which ultimately affects the reward direction during reinforcement learning.

\paragraph{Inverse reward projection.}
For each sampled rollout group, we first cast the reward computed by Eq.~\eqref{eq:delta} and Eq.~\eqref{eq:group_reward} into the Gibbs weights with a proper temperature $T>0$, formulated as follows:
\begin{equation}
\bm r
= \left(r_1, \dots, r_G\right)
= \left(\frac{\exp(R_1(\bm w_{\mathrm{base}})/T)}
{\sum_{i'=1}^{G}\exp(R_{i'}(\bm w_{\mathrm{base}})/T)},\dots,\frac{\exp(R_G(\bm w_{\mathrm{base}})/T)}
{\sum_{i'=1}^{G}\exp(R_{i'}(\bm w_{\mathrm{base}})/T)}  \right).
\label{eq:base_pass}
\end{equation}
Then, for each rollout, we have a reward-based weight $r_i$ corresponding to it. As all rollouts have been evaluated under $K$ criteria,  we can inversely calculate the \emph{average} score of each criterion for these rollouts with $s_{i,j}^{(k)}$, and further estimate the \emph{effective} saturation degree of each criterion with $\bm r$:
\begin{equation}
\bm{\bar s^{(k)}} 
= \left(\bar s^{(k)}_1,\dots, \bar s^{(k)}_G\right)
=\left(\frac{1}{G-1}
\sum_{j\neq 1} s_{1,j}^{(k)},\dots, \frac{1}{G-1}
\sum_{j\neq G} s_{G,j}^{(k)},\right),~
P^{(k)}
=
\frac{\langle \bm r, \bm{\bar s^{(k)}}\rangle}{S_{\max}}.
\label{eq:saturation_estimate}
\end{equation}
Since $\bar s_{i}^{(k)}\in[0,S_{\max}]$ and $\sum_i r_i=1$, this ensures 
$P^{(k)}\in[0,1]$. Based on the above equations, we inversely project the reward back to each criterion to measure the criterion-level saturation under this group of rollouts. Two points are worth noting. First, the saturation degree of each criterion in the rubric is overall positively correlated with the reward. Namely, rollout groups that receive higher rewards tend to exhibit higher criterion saturation. Second, the saturation degree of each criterion may not keep the same pace along with the reward improvement. That is to say, as the RL training progresses and reward improves, different criteria can be satisfied to varying degrees and even become very imbalanced. Such a challenge can be explicitly captured by our quantity $\bm P = (P^{(1)},\dots,P^{(K)})$.

\begin{figure*}[t!]
    \centering
    \includegraphics[width=\textwidth]{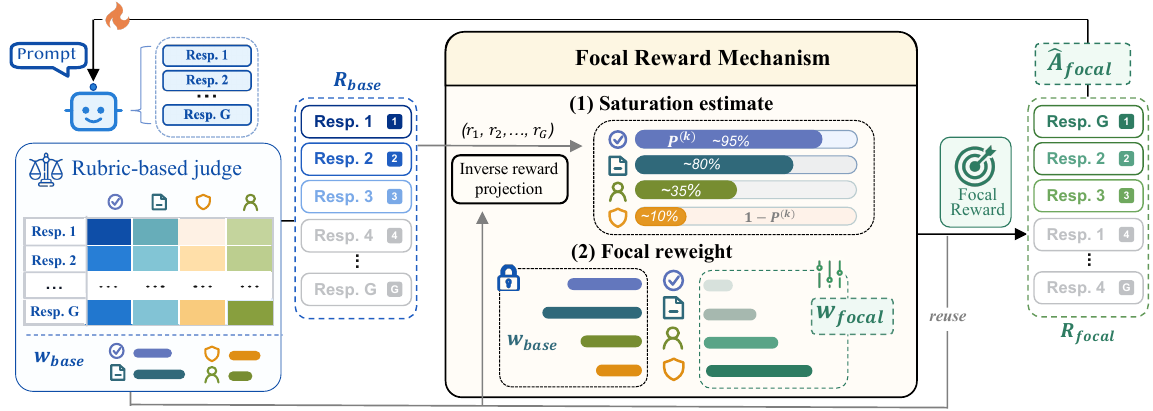}
    \expcaptionsetup
    \caption{
    Overview of \emph{Focal Reward}: it estimates criterion saturation from the base reward, reweights the base scalarizer toward high-headroom criteria, and synthesizes the focal reward for policy optimization. The resulting reward may reorder rollouts and assign higher credit to responses that better satisfy under-saturated criteria.
    }
    \label{fig:focal_reward_method}
    \expfigvspace
\end{figure*}

\paragraph{Focal reweighting aggregation.}
With the quantity $\bm P$ to characterize the saturation of all criteria, Focal Reward modulates the
base rubric static weights $\bm w_\mathrm{base}$ with a headroom-dependent factor $\bm 1-\bm P$. Specifically, we design the focal weights for all criteria in the rubric as follows
\begin{equation}
\bm w_{\mathrm{focal}}
=
\frac{\langle \bm w_\mathrm{base}, \bm 1 \rangle }
{\langle \bm{\tilde w}, \bm 1 \rangle} \bm{\tilde w}
\quad \text{where} \quad
\bm{\tilde w}
= (\bm 1 - \bm P + \epsilon)^\gamma \odot \bm w_{\mathrm{base}}.
\label{eq:focal_weight}
\end{equation}
Here, $\epsilon>0$ is a smoothing constant and $\gamma>0$ controls the strength of
reallocation toward less saturated criteria. Since $\bm 1-\bm P$ estimates remaining
frontier headroom, criteria that remain weak on the current strong-rollout frontier
receive larger multiplicative factors. The normalization preserves the total weight mass of the base scalarizer, so focal reweighting changes relative criterion emphasis without trivially changing the overall reward scale. 
With $\bm w_{\mathrm{focal}}$, we form the following final reward for reinforcement learning, which is then converted to the advantage and used further in Eq.~\eqref{eq:grpo_obj} 
\begin{equation}
R_i(\bm w_{\mathrm{focal}})
=
\sum_{j\neq i}
\Phi_\tau\!\left(\Delta_{i,j}(\bm w_{\mathrm{focal}})\right).
\label{eq:focal_reward}
\end{equation}
Here, the margins $\Delta_{i,j}(\bm w_{\mathrm{focal}})$ are similarly computed based on Eq.~\eqref{eq:delta} with the same criterion-level pairwise
scores $s_{i,j}^{(k)}$ obtained in the base pass. Thus, considering the computation in above equations, the additional cost of Focal Reward is negligible. In Fig.~\ref{fig:focal_reward_method}, we illustrate its whole procedure.

\subsection{Theoretical Analysis}
\label{sec:focal_theory}

In this part, we provide theoretical analysis about transforming the pairwise comparison scores into a scalar reward when the criteria retain different saturation degrees. Due to the space limitation, all assumptions, derivations, and boundary conditions are deferred to Appendix~\ref{app:focal_theory}.


\begin{theorem}[\textbf{Local misallocation bound}]
\label{thm:local_misallocation}
Given a training state at timestep $t$, let $\bm a\in\mathbb{R}_+^K$ be any nonzero criterion scalarizer, $\bm D_t$ denote any criterion-level pairwise score-difference vector, $\bm a^\top \bm D_t$ be the scalarized pairwise margin, and $L_t\in\{+1,-1\}$ be the latent locally useful update direction. Under the local latent-label assumptions in Appendix~\ref{app:focal_theory}, we have local misallocation bound
\[
\Pr\!\left(L_t \bm a^\top \bm D_t\le 0\right)
\le
\exp\!\left(-\frac{1}{2}\Xi_t(\bm a)\right),
\]
where $\Xi_t(\bm a)=\frac{(\bm a^\top\bm \eta_t)^2}{\max_{\ell\in\{+1,-1\}}\bm a^\top\bm\Sigma_t(\ell)\bm a}$ and $\bm\eta_t$ denotes the criterion-level learning signal, $\ell$ is the latent local label, and $\bm \Sigma_t(\ell)$ captures residual criterion-level fluctuation under label $\ell$.
\end{theorem}

Theorem~\ref{thm:local_misallocation} shows that maximizing $\Xi_t(\bm a)$ will reduce local pairwise credit misallocation, where $\Xi_t(\bm a)$ can be interpreted as a local signal-to-noise ratio: the numerator favors scalarizers aligned with useful criterion-wise learning signal, while the denominator penalizes scalarizers whose margins are dominated by residual criterion-level variation.

\begin{theorem}[\textbf{Static aggregation gap}]
\label{thm:static_gap_direction}
Given $\Xi_t(\bm a)$ in Theorem~\ref{thm:local_misallocation}, let $\bm a_t^\star$ denote the locally preferred scalarizer that maximizes $\Xi_t(\bm a)$, and $H_t^{(k)}\in[0,1]$ denote the remaining frontier headroom of criterion $c^{(k)}$ at the timestep $t$. 
Under the symmetric and isotropic simplifications and the headroom-shaped edge model in Appendix~\ref{app:focal_theory}, the preferred criterion scalarizer has the following closed-form direction, and the fixed base scalarizer $\bm w_{\mathrm{base}}$ incurs the corresponding local surrogate gap:
\[
[\bm a_t^\star]^{(k)}
\propto
\bm w_{\mathrm{base}}^{(k)}\bigl(H_t^{(k)}\bigr)^{\gamma_0}, 
\quad
\Xi_t(\bm a_t^\star)-\Xi_t(\bm w_{\mathrm{base}})
=
\frac{c_t^2\|\bm w_{\mathrm{base}}\|_2^2}{\sigma_t^2}\,
\mathrm{Var}_{\mu}\!\left(H_t^{\gamma_0}\right),
\]
where $\gamma_0>0$ is a curvature parameter, $\mu^{(k)}=\frac{\bigl(\bm w_{\mathrm{base}}^{(k)}\bigr)^2}{\|\bm w_{\mathrm{base}}\|_2^2}$,  $\mathrm{Var}_{\mu}(\cdot)$ denotes weighted variance under $\mu$, $c_t>0$ is the state-dependent scale, and $\sigma_t^2$ is the isotropic residual variance scale.
\end{theorem}

Theorem~\ref{thm:static_gap_direction} shows that homogeneous base-weighted headroom makes the variance term vanish, so \(\bm w_{\mathrm{base}}\) remains locally aligned with the preferred direction. When headroom is heterogeneous, the positive variance term indicates that \(\bm w_{\mathrm{base}}\) misses the frontier-headroom factor. This corresponds to our design: Eq.~\eqref{eq:focal_weight} uses \(1-P^{(k)}\) as a proxy for this latent factor, while Appendix~\ref{app:proxy_bridge} provides the proxy bridge by showing that \(P^{(k)}\) estimates frontier-level saturation under Gibbs concentration.

\section{Experiments}
\subsection{Experimental Setup}
\label{sec:exp_setup}

\textbf{Backbone models and Baselines.} We employ diverse LLMs of different scales in our experiments, including Qwen2.5 series (7B-Instruct)~\citep{DBLP:journals/corr/abs-2412-15115} and Qwen3 series (Qwen3-8B/30B-A3B)~\citep{yang2025qwen3}. Beyond the vanilla model, we compare Focal Reward against several static scalarization baselines, including two rule-based baselines: \textbf{Veto}~\citep{DBLP:journals/corr/abs-2508-12790}, using hard-rule-gated aggregation and \textbf{Min-Score}~\citep{pmlr-v235-park24b}, using bottleneck-style weights with minimum criterion score, and two preset baselines: \textbf{Static (uniform)}~\citep{DBLP:conf/icml/CuiY0YH0NXXL0024}, using uniform weights and \textbf{Static (prior)}~\citep{DBLP:journals/corr/abs-2507-18624,DBLP:journals/corr/abs-2507-17746}, using GPT-4o-predicted weights. We apply \textbf{Focal Reward} to both preset weights, obtaining the \textbf{uniform} and \textbf{prior} versions.


\textbf{Implementation Details.} We implement our methods based on the verl framework~\citep{DBLP:conf/eurosys/ShengZYWZZPL025} and the GSPO algorithm~\citep{zheng2025group}, conducted on 32$\times$H800-80GB GPUs. For main experiments, we use two 10K-example OpenRubrics subsets~\citep{DBLP:journals/corr/abs-2510-07743}, from general and science domains separately. During RL training, we use a global batch size of 64 and adopt AdamW with a learning rate of $1\times10^{-6}$. We set the number of rollouts $G=8$, hyper-parameters $\gamma = 2$ and $T=10$. During evaluation, general-domain models are evaluated on AlpacaEval 2.0~\citep{DBLP:journals/corr/abs-2404-04475}, Arena Hard~\citep{DBLP:conf/icml/LiCFD0ZGS25}, WritingBench~\citep{DBLP:journals/corr/abs-2503-05244}, and EQ-Bench 3~\citep{DBLP:journals/corr/abs-2312-06281}; science-domain models are evaluated on GPQA Diamond~\citep{rein2024gpqa} and HealthBench~\citep{DBLP:journals/corr/abs-2505-08775}. See Appendix~\ref{app:training_details} for more training details and benchmark descriptions.

\begin{table*}[t]
\centering
\footnotesize
\expcaptionsetup
\setlength{\tabcolsep}{2.8pt}
\renewcommand{\arraystretch}{0.98}
\caption{\textbf{Main results (\%) on general-domain and science-oriented benchmarks across different backbones.} The first row of each backbone block reports the vanilla model. AE2, AH, WB, EQB3, GPQA, and HB denote AlpacaEval 2.0, Arena-Hard, WritingBench, EQ-Bench 3, GPQA Diamond, and HealthBench, respectively. Bold and underlined scores mark the best and second-best results within each backbone--benchmark block.}
\label{tab:main_results}
\begin{tabular*}{0.90\textwidth}{@{\extracolsep{\fill}}lccccccc@{}}
\toprule
 & \multicolumn{4}{c}{\textbf{General-domain}} & \multicolumn{2}{c}{\textbf{Sci-oriented}} & \multirow{2}{*}{\textbf{Avg.}} \\
\cmidrule(lr){2-5} \cmidrule(lr){6-7}
\multicolumn{1}{c}{Model} & AE2 & AH & WB & EQB3 & GPQA & HB & \\
\midrule

Qwen2.5-7B-Instruct  & 24.50 & 39.91 & 47.16 & 35.31 & 33.74 & 22.29 & 33.82 \\
\quad + Veto & 35.84 & 40.70 & 50.92 & 40.42 & 34.96 & 38.27 & 40.19 \\
\quad + Min-Score & 40.31 & 37.18 & 50.81 & 39.16 & 35.90 & 40.18 & 40.59 \\
\quad + Static (uniform) & 41.22 & 48.06 & 53.52 & 41.78 & 36.05 & 40.31 & 43.49 \\
\quad + Static (prior) & 40.86 & 47.48 & 52.71 & 41.24 & 36.55 & 40.65 & 43.25 \\
\focalrowbg \quad + Focal Reward (uniform) & \textbf{44.06} & \textbf{51.93} & \textbf{54.76} & \textbf{43.72} & \underline{37.44} & \textbf{42.16} & \textbf{45.68} \\
\focalrowbg \quad + Focal Reward (prior) & \underline{43.75} & \underline{50.83} & \underline{53.89} & \underline{43.53} & \textbf{37.63} & \underline{41.89} & \underline{45.25} \\
\midrule

Qwen3-8B  & 53.86 & 49.90 & 63.51 & 47.70 & 48.33 & 36.57 & 49.98 \\
\quad + Veto & 59.41 & 52.47 & 65.41 & 50.49 & 50.23 & 44.20 & 53.70 \\
\quad + Min-Score & 56.83 & 56.24 & 64.97 & 49.68 & 50.08 & 45.93 & 53.96 \\
\quad + Static (uniform) & 69.94 & 60.12 & 67.61 & 52.37 & 51.06 & 46.98 & 58.01 \\
\quad + Static (prior) & 71.81 & \underline{60.23} & 68.04 & 53.07 & 51.31 & 47.37 & 58.64 \\
\focalrowbg \quad + Focal Reward (uniform) & \textbf{74.15} & 59.83 & \textbf{70.50} & \textbf{54.69} & \underline{52.15} & \textbf{48.81} & \textbf{60.02} \\
\focalrowbg \quad + Focal Reward (prior) & \underline{73.96} & \textbf{60.47} & \underline{70.24} & \underline{54.38} & \textbf{52.21} & \underline{48.50} & \underline{59.96} \\
\midrule

Qwen3-30B-A3B  & 63.36 & 55.26 & 64.76 & 51.80 & 51.10 & 40.26 & 54.42 \\
\quad + Veto & 57.49 & 74.80 & 69.79 & 56.42 & 55.43 & 54.38 & 61.38 \\
\quad + Min-Score & 61.12 & 72.19 & 70.31 & 55.20 & 54.47 & 54.53 & 61.30 \\
\quad + Static (uniform) & 73.61 & 79.16 & 71.77 & 57.26 & 55.74 & 55.29 & 65.47 \\
\quad + Static (prior) & 75.63 & 80.21 & 72.05 & 58.13 & 56.22 & 55.95 & 66.36 \\
\focalrowbg \quad + Focal Reward (uniform) & \underline{76.87} & \underline{83.99} & \textbf{73.92} & \underline{58.98} & \underline{57.04} & \underline{56.67} & \underline{67.91} \\
\focalrowbg \quad + Focal Reward (prior) & \textbf{77.59} & \textbf{88.69} & \underline{73.38} & \textbf{59.03} & \textbf{57.07} & \textbf{57.39} & \textbf{68.86} \\
\bottomrule
\end{tabular*}
\expfigvspace
\end{table*}
\subsection{Main results}
\label{sec:main_results}


\textbf{Performance on benchmarks of general and science domains.}
Table~\ref{tab:main_results} reports the results on both general-domain and science-oriented benchmarks across three backbone models. Focal Reward delivers consistent improvements over static scalarization across all model scales. Compared with the strongest static baseline, it improves the average score by up to +2.19 points on Qwen2.5-7B-Instruct, +1.38 points on Qwen3-8B, and +2.50 points on Qwen3-30B-A3B. The gains are observed on both general and science-oriented benchmarks, indicating that Focal Reward synthesis provides a broadly effective training signal rather than benefiting only a particular domain or backbone. In addition, the two Focal Reward variants based on uniform and prior weights perform closely to each other while both surpassing existing reward-synthesis baselines, showing that Focal Reward is not strongly dependent on the specific choice of base weights and can robustly improve different priors.

\textbf{Performance on criteria of different difficulty levels.} Table~\ref{tab:rubric_profile} further analyzes the trained models at the rubric level, comparing Static and Focal Reward across criteria with different difficulty levels. Following Fig.~\ref{fig:weight_level}(a), we first use the base model to perform inference on the training prompts and bucket criteria according to their initial score distribution: criteria with scores below 5 are treated as hard criteria, those above 8.5 as easy criteria, and the remaining ones as medium criteria. Compared with Static scalarization, Focal Reward consistently achieves higher average scores and pass rates on hard criteria, showing that it effectively emphasizes criteria that are initially under-satisfied. The gains also extend to medium criteria, indicating that the adaptive reweighting also improves a broader range of rubric aspects. 
On easy criteria, Focal Reward performs comparably to Static while still maintaining a clear improvement over the base model, indicating that hard-criterion gains do not come at the expense of already-satisfied dimensions. Moreover, Focal Reward produces noticeably shorter rollouts than Static across backbones and domains, suggesting more efficient correction of under-satisfied criteria rather than merely relying on longer, polished generation to gain reward.

\begin{table*}[t]
\centering
\footnotesize
\expcaptionsetup
\setlength{\tabcolsep}{6.5pt}
\renewcommand{\arraystretch}{0.98}
\caption{\textbf{Rubric level results across criteria of different difficulty levels.} Criteria are bucketed by Base model scores following Fig.~\ref{fig:weight_level}(a). Each block includes the Base model inference and results of Static and Focal after training. Pass rate measures the percentage of rollouts whose scores exceed the corresponding threshold.}
\label{tab:rubric_profile}

\begin{tabular}{@{}lllrrrrrrr@{}}
\toprule
\multirow{2}{*}{Domain}
& \multirow{2}{*}{Model}
& \multirow{2}{*}{Method}
& \multirow{2}{*}{Len.}
& \multicolumn{4}{c}{Mean score}
& \multicolumn{2}{c}{Pass rate} \\
\cmidrule(lr){5-8}\cmidrule(lr){9-10}
& & & & All & Hard & Med. & Easy & Hard & Easy \\
\midrule

\multirow{9}{*}{General}
& \multirow{3}{*}{Qwen2.5-7B}
& \bcell{Base}   & \bcell{463}  & \bcell{8.21} & \bcell{3.24} & \bcell{7.65} & \bcell{9.37} & \bcell{-} & \bcell{-} \\
& & Static & 1351 & 8.48 & 5.24 & 7.92 & 9.51 & 39.2 & 88.6 \\
& & \cellcolor{focalgreen!50}Focal & \cellcolor{focalgreen!50}1083 & \cellcolor{focalgreen!50}8.52 & \cellcolor{focalgreen!50}5.71 & \cellcolor{focalgreen!50}7.93 & \cellcolor{focalgreen!50}9.50 & \cellcolor{focalgreen!50}47.4 & \cellcolor{focalgreen!50}88.2 \\

\cmidrule(lr){2-10}
& \multirow{3}{*}{Qwen3-8B}
& \bcell{Base}   & \bcell{658}  & \bcell{8.33} & \bcell{2.87} & \bcell{7.67} & \bcell{9.31} & \bcell{-} & \bcell{-} \\
& & Static & 2446 & 8.60 & 4.82 & 7.99 & 9.45 & 35.9 & 87.9 \\
& & \cellcolor{focalgreen!50}Focal & \cellcolor{focalgreen!50}1806 & \cellcolor{focalgreen!50}8.64 & \cellcolor{focalgreen!50}5.18 & \cellcolor{focalgreen!50}8.02 & \cellcolor{focalgreen!50}9.44 & \cellcolor{focalgreen!50}42.6 & \cellcolor{focalgreen!50}87.6 \\

\cmidrule(lr){2-10}
& \multirow{3}{*}{Qwen3-30B-A3B}
& \bcell{Base}   & \bcell{681}  & \bcell{8.42} & \bcell{2.74} & \bcell{7.69} & \bcell{9.32} & \bcell{-} & \bcell{-} \\
& & Static & 3473 & 8.69 & 4.27 & 7.98 & 9.46 & 29.4 & 88.1 \\
& & \cellcolor{focalgreen!50}Focal & \cellcolor{focalgreen!50}2348 & \cellcolor{focalgreen!50}8.70 & \cellcolor{focalgreen!50}4.56 & \cellcolor{focalgreen!50}7.94 & \cellcolor{focalgreen!50}9.45 & \cellcolor{focalgreen!50}34.5 & \cellcolor{focalgreen!50}87.8 \\

\midrule

\multirow{9}{*}{Science}
& \multirow{3}{*}{Qwen2.5-7B}
& \bcell{Base}   & \bcell{647}  & \bcell{8.11} & \bcell{3.68} & \bcell{7.59} & \bcell{9.34} & \bcell{-} & \bcell{-} \\
& & Static & 3439 & 8.51 & 6.22 & 8.11 & 9.48 & 39.0 & 89.3 \\
& & \cellcolor{focalgreen!50}Focal & \cellcolor{focalgreen!50}2836 & \cellcolor{focalgreen!50}8.55 & \cellcolor{focalgreen!50}6.63 & \cellcolor{focalgreen!50}8.13 & \cellcolor{focalgreen!50}9.49 & \cellcolor{focalgreen!50}49.4 & \cellcolor{focalgreen!50}90.1 \\

\cmidrule(lr){2-10}
& \multirow{3}{*}{Qwen3-8B}
& \bcell{Base}   & \bcell{1055} & \bcell{8.41} & \bcell{2.58} & \bcell{7.69} & \bcell{9.25} & \bcell{-} & \bcell{-} \\
& & Static & 4992 & 8.70 & 4.92 & 8.11 & 9.39 & 41.6 & 88.4 \\
& & \cellcolor{focalgreen!50}Focal & \cellcolor{focalgreen!50}4131 & \cellcolor{focalgreen!50}8.74 & \cellcolor{focalgreen!50}5.34 & \cellcolor{focalgreen!50}8.15 & \cellcolor{focalgreen!50}9.40 & \cellcolor{focalgreen!50}49.2 & \cellcolor{focalgreen!50}89.0 \\

\cmidrule(lr){2-10}
& \multirow{3}{*}{Qwen3-30B-A3B}
& \bcell{Base}   & \bcell{1020} & \bcell{8.67} & \bcell{2.47} & \bcell{7.76} & \bcell{9.28} & \bcell{-} & \bcell{-} \\
& & Static & 6120 & 8.84 & 4.85 & 7.89 & 9.42 & 38.4 & 88.7 \\
& & \cellcolor{focalgreen!50}Focal & \cellcolor{focalgreen!50}5000 & \cellcolor{focalgreen!50}8.88 & \cellcolor{focalgreen!50}5.18 & \cellcolor{focalgreen!50}7.95 & \cellcolor{focalgreen!50}9.43 & \cellcolor{focalgreen!50}47.5 & \cellcolor{focalgreen!50}89.2 \\

\bottomrule
\end{tabular}
\expfigvspace
\end{table*}

\subsection{Further analysis}
\label{sec:mechanism_validation}

While Tables~\ref{tab:main_results} and \ref{tab:rubric_profile} demonstrate the superior performance of Focal Reward, we further analyze the mechanisms underlying these gains through answering the following research questions: 1) How does Focal Reward reallocate weights across rubric criteria? 2) How does this reallocation reshape the resulting reward distribution? 3) Does the model exhibit measurable improvements on the targeted rubric criteria? 4) How sensitive is Focal Reward to the choice of prior weights?


\begin{figure*}[t]
    \centering
    \expcaptionsetup
    \includegraphics[width=\linewidth]{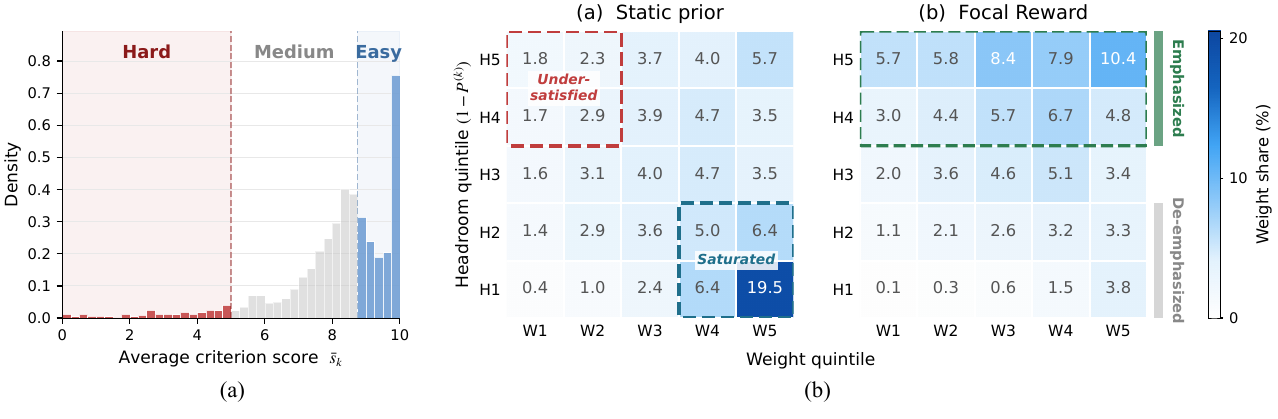}
    \caption{\textbf{Criterion saturation and weight allocation.} (a) Distribution of average criterion scores $\bar{s}^{(k)}$, based on which criteria are divided into hard, medium and easy groups. (b) Relationship between criterion weights and remaining headroom. Criteria are grouped into quintiles according to their assigned weights and estimated headroom, each heatmap reports the corresponding weight share.}
    \label{fig:weight_level}
    \expfigvspace
\end{figure*}

\begin{figure*}[t]
    \centering
    \expcaptionsetup
    \includegraphics[width=\linewidth]{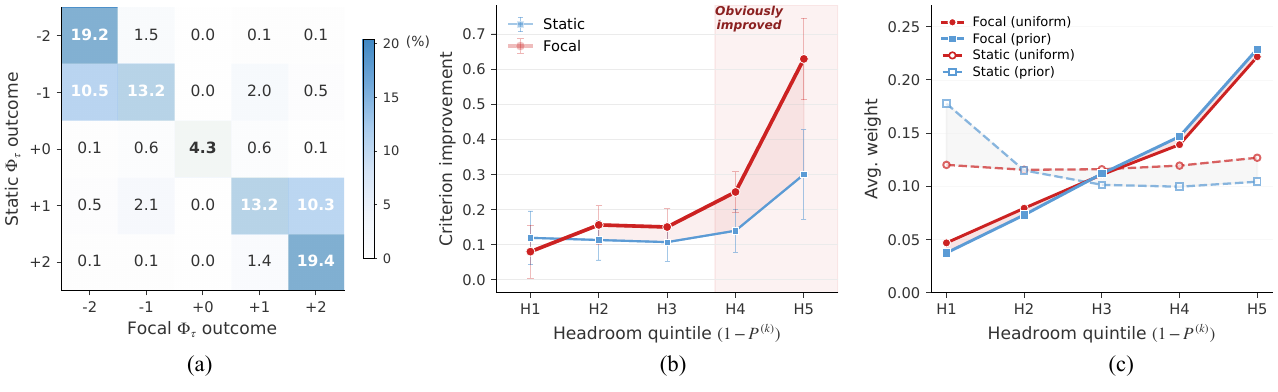}
    \caption{\textbf{Mechanism analysis of Focal Reward.} (a) Transition heatmap of rollout-level reward outcomes under Static and Focal Reward. Each cell reports the percentage of samples moving from Static to Focal outcomes. (b) Improvement on criteria of different estimated remaining headroom, which is measured by the score gain over the untrained base model. (c) Average weight across headroom quintiles for Uniform and Prior initialization before and after focal reweighting. Experiments are conducted on Qwen2.5-7B-Instruct under general domain.}
    \label{fig:reward_level}
    \expfigvspace
\end{figure*}

\textbf{Weight aspect: More emphasis on under-satisfied criterion.} Figure~\ref{fig:weight_level}(b) compares the relationship between criterion weights and estimated remaining headroom under Static prior and Focal Reward. Higher y-axis values indicate more under-satisfied criteria, while higher x-axis values indicate larger weights. Under Static prior, many high-headroom criteria still receive small weights, and some low-headroom criteria remain heavily emphasized, showing that fixed weights fail to track criterion saturation. In contrast, Focal Reward shifts more weight mass toward high-headroom criteria, showing that the adaptive weights can refocus training on criteria with substantial room for improvement.


\textbf{Reward aspect: Better discrimination ability from high-headroom criterion.} Figure~\ref{fig:reward_level}(a) analyzes how rollout-level reward outcomes change after focal reweighting. Focal Reward largely preserves the behavior of Static prior, with around 70\% of samples keeping the same outcome (remaining on the diagonal), indicating a similar overall reward style. Meanwhile, the adjusted rewards become more polarized: roughly 10\% of samples move from $+1$ to $+2$, and another 10\% from $-1$ to $-2$. This suggests that by assigning greater emphasis to under-satisfied criteria, Focal Reward makes these criterion differences more visible and improves the discrimination ability of the reward signal.

\textbf{Optimization aspect: Obvious improvement on targeted criteria.} Figure~\ref{fig:reward_level}(b) validates whether the criteria emphasized by Focal Reward are indeed improved during training. We divide criteria into quintiles according to their estimated remaining headroom $1-P^{(k)}$ and measure the improvement of each criterion as the score gain over the base model. Static scalarization brings only mild improvements on high-headroom criteria, suggesting that fixed weights are insufficient to fully exploit these under-satisfied criteria. In contrast, Focal Reward shows a clear monotonic trend: criteria with larger remaining headroom obtain larger improvements, with the highest-headroom group exhibiting the strongest gain. This confirms that the proposed reweighting mechanism successfully converts headroom-aware reward correction into targeted rubric-level optimization.

\textbf{Prior aspect: Robustness across various base scalarizers.} Figure~\ref{fig:reward_level}(c) compares how different base scalarizers evolve after focal reweighting. The base weights encode the judge's prior emphasis over criteria and can therefore influence both rollout rewards and downstream performance. As depicted, Uniform and Prior initialization exhibit clearly different weight distributions across headroom levels. Interestingly, despite starting from these distinct priors, Focal Reward drives the weights toward highly similar distributions, with a cosine similarity of about $0.9$. Both variants consistently allocate larger weights to high-headroom criteria, showing the robustness of Focal Reward to base weights.

\begin{table*}[t]
    \centering
    \expcaptionsetup
    \begin{minipage}[t]{0.50\textwidth}
    \centering
    \vspace{0pt}
    \scriptsize
    \setlength{\tabcolsep}{2.4pt}
    \renewcommand{\arraystretch}{0.92}
    \caption{\textbf{Component ablation of Focal Reward} on Qwen2.5-7B-Instruct under general domain. \textit{Focal w/o frontier weighting} ablates the frontier-aware saturation by averaging uniformly over rollouts. \textit{Focal w/ frozen scalarizer} ablates online adaptation by keeping the initially computed focal weights fixed throughout training.}
    \label{tab:ablation}
    \begin{adjustbox}{max width=\linewidth}
    \begin{tabular}{lccccc}
    \toprule
    Method & AE2 & AH & WB & EQB3 & Avg. \\
    \midrule
    Static  & 41.22 & 48.06 & 53.52 & 41.78 & 46.15 \\
    Focal Reward  & \textbf{44.06} & \textbf{51.93} & \textbf{54.76} & \textbf{43.72} & \textbf{48.62} \\
    \midrule
    Focal w/o frontier weighting & 41.75 & 48.73 & 53.96 & 42.35 & 46.70 \\
    Focal w/ frozen scalarizer & 40.97 & 47.67 & 53.63 & 41.89 & 46.04 \\
    \bottomrule
    \end{tabular}
    \end{adjustbox}
    \end{minipage}
    \hfill
    \begin{minipage}[t]{0.47\textwidth}
    \centering
    \vspace{0pt}
    \scriptsize
    \setlength{\tabcolsep}{4pt}
    \renewcommand{\arraystretch}{0.92}
    \caption{\textbf{Runtime overhead of Focal Reward.} Results are averaged over multiple training runs with Qwen2.5-7B-Instruct backbone on the general-domain dataset. \textit{Len.} denotes the average rollout length. \textit{Synth.} and \textit{Rollout} denote the reward-synthesis and rollout generating time separately. And \textit{Total Step} denotes the end-to-end training step time.}
    \label{tab:runtime}
    \begin{adjustbox}{max width=\linewidth}
    \begin{tabular}{lrrrr}
    \toprule
    Method & Len. & Synth. & Rollout & Total Step \\
    \midrule
    Static       & 1351 & 14 ms  & 225.8 s & 282.0 s \\
    Focal Reward & 1083 & 50 ms  & 153.9 s & 175.4 s \\
    \midrule
    Abs. $\Delta$ & $-$268 & +36 ms & $-$71.9 s & $-$106.6 s \\
    \bottomrule
    \end{tabular}
    \end{adjustbox}
    \end{minipage}
    \expfigvspace
\end{table*}



\subsection{Ablation studies}
\label{sec:ablation}

\begin{wrapfigure}[11]{r}{0.455\linewidth}
    \expfigvspace
    \centering
    \captionsetup{font=small,skip=2pt}
    \includegraphics[width=\linewidth]{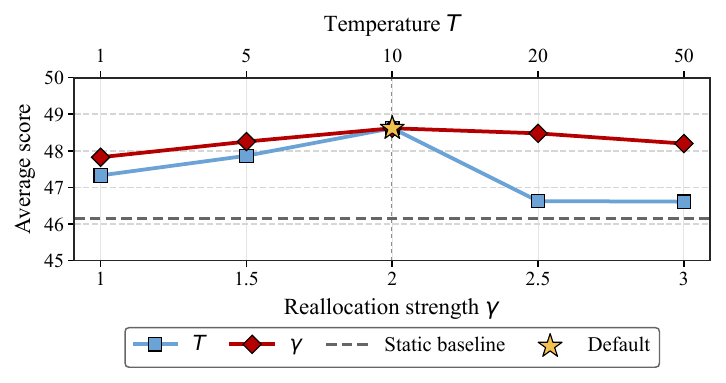}
    \caption{\textbf{Sensitivity analysis of hyper-parameters $\gamma$ and $T$} on Qwen2.5-7B-Instruct in general-domain. The star marks the default setting.}
    \label{fig:hyperparameter_sensitivity}
    \expfigvspace
\end{wrapfigure}

\begin{figure}[t]
    \expcaptionsetup
    \centering
    \captionsetup{font=small,skip=2pt}
    \includegraphics[width=\linewidth]{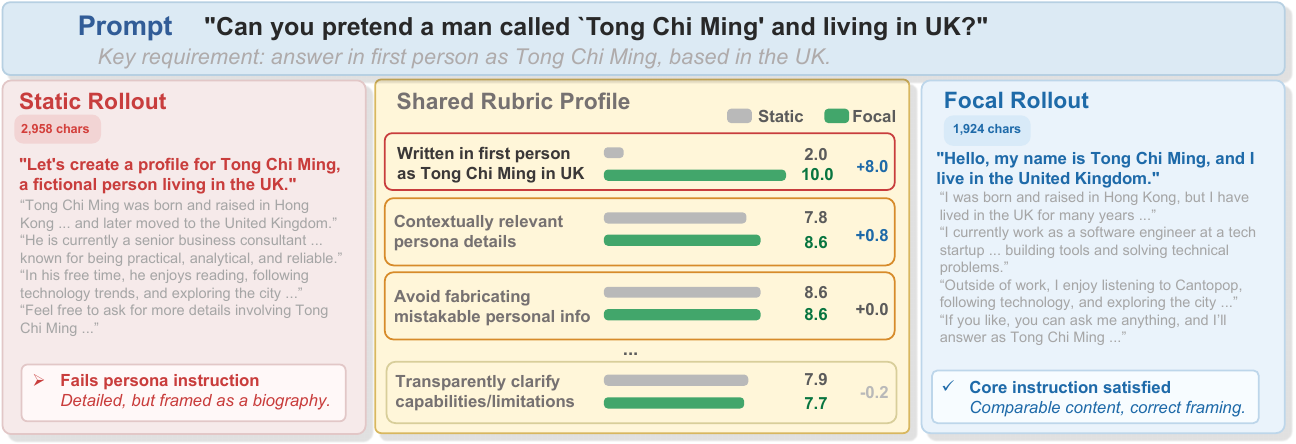}
    \caption{\textbf{Qualitative case study.} Static misses the first-person role-play requirement despite detailed content, while Focal preserves rollout quality and corrects this weakness by emphasizing the under-satisfied hard rule.}
    \label{fig:case_study}
    \expfigvspace
\end{figure}

\textbf{Component ablation of Focal Reward.} Table~\ref{tab:ablation} studies the contribution of the key components in Focal Reward. Removing frontier weighting drops the average score from 48.62 to 46.70, showing that saturation should be estimated from the strong-rollout frontier rather than uniformly over all rollouts. Freezing the scalarizer further reduces the score to 46.04, indicating that a fixed reward direction cannot track the changing saturation states during training. These results suggest that both the frontier-aware estimation and the online reweighting components are essential for effective reward reallocation.

\textbf{Computational cost analysis.} Table~\ref{tab:runtime} shows that Focal Reward remains computationally efficient compared with the Static baseline. Since it reuses the same rubric-level judgment and only applies lightweight reweighting during reward synthesis, it introduces negligible computational overhead. Moreover, the model optimized with Focal Reward produces shorter rollouts while maintaining overall quality, which further reduces generation cost and leads to lower end-to-end training overhead.

\textbf{Hyper-parameter sensitivity analysis.} Figure~\ref{fig:hyperparameter_sensitivity} analyzes the sensitivity of Focal Reward to the reallocation strength $\gamma$ and temperature $T$. Across a wide range of values, Focal Reward consistently outperforms the Static baseline, indicating that its gains are not tied to a narrow hyper-parameter choice. For $\gamma$, the performance remains stable around the default setting. For $T$, the performance is also stable within the range of 1 to 10, while an overly large temperature may over-smooth the reward-based rollout weights and weaken the saturation estimate, leading to a slight performance drop. Overall, Focal Reward shows robustness under different hyper-parameter configurations.

\FloatBarrier

\textbf{Case study.}
Figure~\ref{fig:case_study} provides a qualitative example of the failure mode discussed in Figure~\ref{fig:intro_overview}(c). The prompt requires first-person role play, yet Static ranks a fluent and detailed rollout highly because it performs well on broad quality criteria while missing the decisive task-specific requirement, namely the correct first-person framing. In contrast, Focal Reward assigns greater emphasis to this under-satisfied criterion and favors a rollout that follows the intended role-play format. This example qualitatively illustrates the central mechanism of Focal Reward: reallocating reward emphasis toward criteria with substantial remaining headroom can correct high-reward but behaviorally flawed generations. More examples and full rollouts are provided in Appendix~\ref{app:case_study}.

\section{Related Work}
\label{sec:related}

\noindent\textbf{Rubric-based RLAIF.}
Open-ended generation is subjective and multi-dimensional, so modern alignment often uses rubric-based AI judges to provide structured, criterion-level feedback rather than holistic or opaque preference signals \citep{DBLP:conf/icml/0001PMMFLBHCRP24,DBLP:journals/corr/abs-2507-17746,DBLP:journals/corr/abs-2507-18624}. Prior work has explored rubric-based rewards, checklists, scalable rubric generation, context-aware criteria, adaptive pairwise rubrics, and adaptive rubric learning \citep{DBLP:journals/corr/abs-2508-12790,DBLP:journals/corr/abs-2510-07743,DBLP:conf/acl/GuptaSZMGBYR25,DBLP:journals/corr/abs-2602-14069,DBLP:journals/corr/abs-2602-20751}. These studies mainly improve how rubrics are constructed, adapted, or judged. We study the reward-synthesis step under a fixed rubric and judge, asking how criterion-level feedback should be aggregated into scalar rewards for policy optimization.

\noindent\textbf{Dynamic Reward Weighting.}
Reward aggregation in rubric-based reinforcement learning can be viewed as a scalarization problem, where criterion-level feedback is compressed into a scalar reward for policy optimization. Prior work has studied scalarization and preference trade-offs in broader multi-objective alignment \citep{DBLP:conf/emnlp/00030X0024,DBLP:conf/acl/ZhouLS00O024,DBLP:conf/pkdd/HeM25}. Recent methods further adapt weights according to context or prompt-specific preferences \citep{DBLP:conf/icml/0010PLQ00C24,DBLP:journals/corr/abs-2511-10656}, or according to optimization states, learning potential, and objective conflicts \citep{DBLP:journals/corr/abs-2509-11452,DBLP:journals/corr/abs-2601-06574}. These methods mainly adapt weights across high-level objectives or preference dimensions, leaving the fine-grained criterion-level aggregation problem under a fixed rubric and fixed judge less explored. More related work is discussed in Appendix~\ref{app:related_work}.
\section{Conclusion}


In this work, we introduce Focal Reward, a novel objective for automatically balancing reinforcement learning under rubric-based rewards. It addresses imbalanced reward polarization caused by static scalarizers, which may continue to emphasize already-satisfied criteria while under-emphasizing criteria that still retain substantial room for improvement. Focal Reward estimates criterion-level saturation through inverse reward projection and adaptively reallocates reward emphasis from saturated criteria toward high-headroom criteria, thereby calibrating the reward direction before policy optimization. Extensive experiments and analyses demonstrate its effectiveness, showing improved reward discrimination and targeted rubric-level gains. These findings highlight the importance of adaptive reward synthesis for effective policy optimization in open-ended generation.

\newpage
\bibliographystyle{plainnat}
\bibliography{references}

\newpage
\appendix

\clearpage
\appendix
\startcontents[appendix]

\phantomsection
\section*{Appendix}

\vspace{0.8em}
{\LARGE\bfseries Table of Contents\par}
\vspace{0.8em}

{
\setcounter{tocdepth}{2}
\contentsmargin{0em}

\titlecontents{section}
  [0em]
  {\vspace{0.45em}\normalsize\bfseries\color{blue!50!black}}
  {\contentslabel{2.2em}}
  {}
  {\hfill\textcolor{black}{\contentspage}}

\titlecontents{subsection}
  [2.8em]
  {\vspace{0.05em}\small\color{blue!50!black}}
  {\contentslabel{2.8em}}
  {}
  {\titlerule*[0.55em]{.}\textcolor{black}{\contentspage}}

\printcontents[appendix]{}{1}{}
}

\clearpage

\section{Limitations and Broader Impact}
\label{app:limitations-impact}



\paragraph{Limitations.}
Focal Reward operates on criterion-level judge scores and therefore benefits from well-calibrated rubric feedback, as is common in rubric-based RL. If the judge confuses criteria, exhibits systematic biases, or assigns poorly calibrated absolute scores, the estimated saturation levels may inherit these errors and affect the resulting reweighting. In our experiments, we mitigate this issue by keeping the judge and rubric fixed across all methods, using pairwise criterion-level scoring, and evaluating across multiple benchmarks and model scales. The theory is intended to characterize the local reward-allocation mechanism rather than to provide a full convergence analysis of large-scale RL with non-stationary LLM policies. This scope matches the empirical focus of the paper, where the method is evaluated through rollout-level, mechanism-level, and ablation analyses. Focal Reward adds only lightweight reward-synthesis computation, while the overall RL post-training pipeline remains resource-intensive due to rollout generation and rubric-based judging.

\paragraph{Broader impact.}
Focal Reward may contribute to more balanced, reliable, and human-aligned open-ended language-model systems by encouraging optimization across diverse rubric criteria rather than allowing training to concentrate on already-satisfied dimensions. By reallocating reward emphasis toward under-satisfied criteria, it can improve the overall quality and consistency of model rollouts, especially for hard requirements such as instruction following, safety constraints, factual grounding, domain-specific completeness, and fairness-related criteria when they are explicitly included in the rubric. This may help reduce uneven model behavior in real user-facing scenarios, where a response that is fluent or well-formatted can still fail important requirements. More broadly, saturation-aware reward synthesis provides a practical mechanism for making rubric-based RL more balanced and diagnosable, allowing developers to identify which quality dimensions remain under-optimized and to direct training pressure accordingly. 

At the same time, Focal Reward inherits the limitations and risks of the rubric and judge used to produce criterion level scores. If the rubric omits important safety, privacy, or fairness criteria, or if the judge systematically miscalibrates certain criteria, the adaptive reweighting may amplify these incomplete or biased signals during training. Stronger open ended generation systems may also be misused for misleading, low quality, or harmful content generation when deployed without appropriate safeguards. Practical mitigation strategies include auditing rubrics before training, evaluating judge calibration and bias, explicitly including safety and fairness related criteria, and monitoring downstream model behavior after deployment.
\section{Additional Details of GRPO}
\label{app:details_GRPO}

We follow the GRPO formulation of
\citet{shao2024deepseekmath}. Given a prompt $x\in\mathcal{D}$, the old
policy $\pi_{\theta_{\mathrm{old}}}$ samples a group of $G$ candidate
rollouts $Y=\{y_i\}_{i=1}^{G}\sim\pi_{\theta_{\mathrm{old}}}(\cdot\mid x)$.
After rubric-based judgment and reward aggregation, each rollout $y_i$
receives a scalar reward $R_i$. GRPO estimates the advantage using the
relative rewards within the same sampled group, without learning an
additional value function:
\begin{equation}
\hat A_i
=
\frac{
R_i-\mathrm{mean}\bigl(\{R_1,\ldots,R_G\}\bigr)
}{
\mathrm{std}\bigl(\{R_1,\ldots,R_G\}\bigr)
}.
\label{eq:grpo_advantage_detail}
\end{equation}
Since $R_i$ is assigned at the rollout level, the same advantage $\hat A_i$
is used for all tokens in $y_i$.

For each generated token $y_{i,t}$, the token-level importance ratio is
defined as
\begin{equation}
r_{i,t}(\theta)
=
\frac{
\pi_{\theta}(y_{i,t}\mid x,y_{i,<t})
}{
\pi_{\theta_{\mathrm{old}}}(y_{i,t}\mid x,y_{i,<t})
}.
\label{eq:grpo_ratio_detail}
\end{equation}
The policy is then optimized with the clipped objective in
Eq.~\eqref{eq:grpo_obj}. Following GRPO, the KL penalty is estimated at the
token level as
\begin{equation}
D_{\mathrm{KL}}(\pi_{\theta}\Vert\pi_{\mathrm{ref}})
=
\frac{
\pi_{\mathrm{ref}}(y_{i,t}\mid x,y_{i,<t})
}{
\pi_{\theta}(y_{i,t}\mid x,y_{i,<t})
}
-
\log
\frac{
\pi_{\mathrm{ref}}(y_{i,t}\mid x,y_{i,<t})
}{
\pi_{\theta}(y_{i,t}\mid x,y_{i,<t})
}
-1 .
\label{eq:grpo_kl_detail}
\end{equation}

In our setting, Static aggregation and Focal Reward differ only in how the
criterion-level feedback is synthesized into the scalar rollout reward
$R_i$. Once $R_i$ is obtained, both methods use the same group-relative
normalization in Eq.~\eqref{eq:grpo_advantage_detail}, the same token-level
ratio in Eq.~\eqref{eq:grpo_ratio_detail}, and the same clipped policy
optimization objective in Eq.~\eqref{eq:grpo_obj}. Thus, Focal Reward
changes the reward direction before policy optimization while leaving the
GRPO optimization procedure unchanged.
\section{Additional Theoretical Details for Focal Reward}
\label{app:focal_theory}

This appendix complements Section~\ref{sec:focal_theory}. 
The main text states only the key local conclusions: Theorem~\ref{thm:local_misallocation} gives the pairwise misallocation bound that motivates the surrogate $\Xi_t(\bm a)$, and Theorem~\ref{thm:static_gap_direction} characterizes the static aggregation gap relative to the headroom-aware preferred direction. 
Here we provide the formal local latent-label model, derive the misallocation bound, prove the closed-form preferred direction and static gap, and then discuss the proxy bridge and identification boundary for the implemented saturation statistic.

\subsection{Local latent-label model}
\label{app:pairwise_model}

Given a training state at the timestep $t$, all probabilities and expectations below are taken with respect to the joint law of
\[
X\sim\mathcal D,
\qquad
Y=(Y_1,\ldots,Y_G)\sim \pi_{\theta_t}(\cdot\mid X)^G,
\qquad
(I,J)\sim \mathrm{Unif}\big\{(i,j):1\le i,j\le G,\ i\neq j\big\},
\]
together with a latent label $L_t\in\{+1,-1\}$.

For each ordered pair $(i,j)$, define the criterion-level pairwise score-difference vector
\begin{equation}
\label{eq:app_pairwise_diff}
d_{i,j}^{(k)}:=s_{i,j}^{(k)}-s_{j,i}^{(k)},
\qquad
\bm d_{i,j}:=\bigl(d_{i,j}^{(1)},\ldots,d_{i,j}^{(K)}\bigr)\in\mathbb R^K .
\end{equation}
The random ordered pair induces the random criterion-level score-difference vector
\begin{equation}
\label{eq:app_random_pairwise_vector}
\bm D_t:=\bm d_{I,J}(X,Y)\in\mathbb R^K .
\end{equation}
For any nonnegative scalarizer $\bm a\in\mathbb R_+^K$, the scalarized pairwise margin is given by $\bm a^\top\bm D_t$.

The latent variable $L_t\in\{+1,-1\}$ indicates which direction of pairwise credit assignment is locally more useful for policy improvement at the timestep $t$. The analysis uses the following assumptions.

\paragraph{(A1) Positive learning edge.}
There exists $\bm\eta_t\in\mathbb R_+^K$ such that
\begin{equation}
\label{eq:app_edge_assumption}
\mathbb E[\bm D_t\mid L_t=\ell]=\ell\,\bm\eta_t,
\qquad \ell\in\{+1,-1\}.
\end{equation}

\paragraph{(A2) Headroom-shaped edge.}
There exist latent frontier-saturation levels $q_t^{(k)}\in[0,1]$, latent remaining frontier headroom values $H_t^{(k)}:=1-q_t^{(k)}$, and constants $c_t>0$, $\gamma_0>0$ such that
\begin{equation}
\label{eq:app_headroom_assumption}
\eta_t^{(k)}
=
c_t\,w_{\mathrm{base}}^{(k)}\bigl(H_t^{(k)}\bigr)^{\gamma_0},
\qquad k=1,\ldots,K.
\end{equation}

\paragraph{(A3) Proxy-sub-Gaussian residual.}
Define the residual
\begin{equation}
\label{eq:app_residual}
\bm\zeta_t:=\bm D_t-L_t\bm\eta_t .
\end{equation}
For each $\ell\in\{+1,-1\}$, there exists a symmetric positive-definite matrix $\bm\Sigma_t(\ell)\succ0$ such that
\begin{equation}
\label{eq:app_proxy_subgaussian}
\mathbb E\!\left[\exp\!\bigl(\bm u^\top\bm\zeta_t\bigr)\mid L_t=\ell\right]
\le
\exp\!\left(\frac12 \bm u^\top\bm\Sigma_t(\ell)\bm u\right),
\qquad \forall \bm u\in\mathbb R^K.
\end{equation}
Thus $\bm\Sigma_t(\ell)$ is a conditional sub-Gaussian proxy matrix. It need not equal the true conditional covariance of $\bm\zeta_t$. 
This assumption is used only as a proxy tail condition for deriving the local bound. Since criterion-level scores are bounded in practice, residual fluctuations can be controlled by a sub-Gaussian proxy matrix, which may absorb judge bias, criterion correlation, and prompt-level variability.

\paragraph{Discussion of the assumptions.}
Assumption (A1) introduces a local positive learning edge, which allows us to
define pairwise credit misallocation as the event that the scalarized margin has
a sign inconsistent with \(L_t\). Assumption (A3) controls the residual
fluctuation around this edge and yields the exponential bound in
Theorem~\ref{thm:local_misallocation}. Assumption (A2) is used in
Theorem~\ref{thm:static_gap_direction} to connect the preferred scalarizer direction to
remaining frontier headroom. Together, these assumptions provide a local
justification for reallocating reward weight from saturated criteria toward
criteria that still retain larger headroom.

\subsection{Proof of Theorem~\ref{thm:local_misallocation}: From the misallocation bound to the local surrogate}
\label{app:th1_proof}

We derive the local surrogate used in Section~\ref{sec:focal_theory}. 
Under the local latent-label model in Appendix~\ref{app:pairwise_model}, a scalarizer can misallocate pairwise credit when the scalarized margin $\bm a^\top\bm D_t$ has a sign inconsistent with the latent locally useful direction $L_t$. 
The following proof establishes the exponential bound stated in Theorem~\ref{thm:local_misallocation}, which naturally motivates maximizing the surrogate $\Xi_t(\bm a)$.

\paragraph{Restatement of Theorem~\ref{thm:local_misallocation}.}
Assume (A1) and (A3). For any nonzero $\bm a\in\mathbb R_+^K$,
\begin{equation}
\label{eq:app_main_bound}
\Pr\!\left[L_t\,\bm a^\top\bm D_t\le 0\right]
\le
\exp\!\left(
-\frac{(\bm a^\top\bm\eta_t)^2}
{2\max_{\ell\in\{+1,-1\}}\bm a^\top\bm\Sigma_t(\ell)\bm a}
\right).
\end{equation}
Equivalently,
\begin{equation}
\label{eq:app_main_bound_alt}
\Pr\!\left[L_t\,\bm a^\top\bm D_t\le 0\right]
\le
\exp\!\left(
-\frac12
\min_{\ell\in\{+1,-1\}}
\frac{(\bm a^\top\bm\eta_t)^2}{\bm a^\top\bm\Sigma_t(\ell)\bm a}
\right).
\end{equation}

\paragraph{Proof.}
For brevity, write
\[
\bm D:=\bm D_t,
\qquad
L:=L_t,
\qquad
\bm\eta:=\bm\eta_t,
\qquad
\bm\zeta:=\bm\zeta_t,
\qquad
\bm\Sigma_\ell:=\bm\Sigma_t(\ell),\quad \ell\in\{+1,-1\}.
\]
Fix a nonzero scalarizer $\bm a\in\mathbb R_+^K$. Since $\bm a\in\mathbb R_+^K$ and $\bm\eta\in\mathbb R_+^K$, the scalar
\[
\nu:=\bm a^\top\bm\eta
\]
satisfies $\nu\ge 0$. If $\nu=0$, then the right-hand side of \eqref{eq:app_main_bound} equals $1$, so the claim is trivial because probabilities are at most $1$. We therefore assume $\nu>0$.

First condition on the event $\{L=+1\}$. By (A1),
\[
\mathbb E[\bm D\mid L=+1]=\bm\eta.
\]
Because $\bm\zeta=\bm D-L\bm\eta$, on $\{L=+1\}$ we have
\[
\bm D=\bm\eta+\bm\zeta,
\qquad
\bm a^\top\bm D=\bm a^\top\bm\eta+\bm a^\top\bm\zeta.
\]
Hence,
\[
L\,\bm a^\top\bm D\le 0
\iff
\bm a^\top\bm D\le 0
\iff
\bm a^\top\bm\zeta\le -\nu.
\]
For any $\lambda>0$, Markov's inequality gives
\[
\Pr\!\left[\bm a^\top\bm\zeta\le -\nu\mid L=+1\right]
=
\Pr\!\left[
e^{-\lambda\bm a^\top\bm\zeta}\ge e^{\lambda\nu}
\mid L=+1
\right]
\le
e^{-\lambda\nu}
\mathbb E\!\left[
e^{-\lambda\bm a^\top\bm\zeta}
\mid L=+1
\right].
\]
Applying (A3) with $\bm u=-\lambda\bm a$ yields
\[
\mathbb E\!\left[
e^{-\lambda\bm a^\top\bm\zeta}
\mid L=+1
\right]
\le
\exp\!\left(
\frac12\lambda^2\bm a^\top\bm\Sigma_{+1}\bm a
\right).
\]
Therefore,
\[
\Pr\!\left[
L\,\bm a^\top\bm D\le 0
\mid L=+1
\right]
\le
\exp\!\left(
-\lambda\nu
+
\frac12\lambda^2\bm a^\top\bm\Sigma_{+1}\bm a
\right).
\]
The exponent is minimized over $\lambda>0$ by
\[
\lambda^\star
=
\frac{\nu}{\bm a^\top\bm\Sigma_{+1}\bm a},
\]
where the denominator is positive because $\bm\Sigma_{+1}\succ0$ and $\bm a\neq 0$. Substituting this value gives
\begin{equation}
\label{eq:app_plus_bound}
\Pr\!\left[
L\,\bm a^\top\bm D\le 0
\mid L=+1
\right]
\le
\exp\!\left(
-\frac{\nu^2}{2\bm a^\top\bm\Sigma_{+1}\bm a}
\right)
=
\exp\!\left(
-\frac{(\bm a^\top\bm\eta)^2}{2\bm a^\top\bm\Sigma_{+1}\bm a}
\right).
\end{equation}

Now condition on the event $\{L=-1\}$. By (A1),
\[
\mathbb E[\bm D\mid L=-1]=-\bm\eta.
\]
Because $\bm\zeta=\bm D-L\bm\eta$, on $\{L=-1\}$ we have
\[
\bm D=-\bm\eta+\bm\zeta,
\qquad
\bm a^\top\bm D=-\bm a^\top\bm\eta+\bm a^\top\bm\zeta.
\]
Thus,
\[
L\,\bm a^\top\bm D\le 0
\iff
-\bm a^\top\bm D\le 0
\iff
\bm a^\top\bm D\ge 0
\iff
\bm a^\top\bm\zeta\ge \nu.
\]
Again, for any $\lambda>0$, Markov's inequality gives
\[
\Pr\!\left[
\bm a^\top\bm\zeta\ge \nu
\mid L=-1
\right]
=
\Pr\!\left[
e^{\lambda\bm a^\top\bm\zeta}\ge e^{\lambda\nu}
\mid L=-1
\right]
\le
e^{-\lambda\nu}
\mathbb E\!\left[
e^{\lambda\bm a^\top\bm\zeta}
\mid L=-1
\right].
\]
Applying (A3) with $\bm u=\lambda\bm a$ yields
\[
\mathbb E\!\left[
e^{\lambda\bm a^\top\bm\zeta}
\mid L=-1
\right]
\le
\exp\!\left(
\frac12\lambda^2\bm a^\top\bm\Sigma_{-1}\bm a
\right).
\]
Therefore,
\[
\Pr\!\left[
L\,\bm a^\top\bm D\le 0
\mid L=-1
\right]
\le
\exp\!\left(
-\lambda\nu
+
\frac12\lambda^2\bm a^\top\bm\Sigma_{-1}\bm a
\right).
\]
Optimizing over $\lambda>0$ gives
\[
\lambda^\star
=
\frac{\nu}{\bm a^\top\bm\Sigma_{-1}\bm a},
\]
and therefore
\begin{equation}
\label{eq:app_minus_bound}
\Pr\!\left[
L\,\bm a^\top\bm D\le 0
\mid L=-1
\right]
\le
\exp\!\left(
-\frac{(\bm a^\top\bm\eta)^2}{2\bm a^\top\bm\Sigma_{-1}\bm a}
\right).
\end{equation}

Finally, by the law of total probability,
\[
\Pr\!\left[
L\,\bm a^\top\bm D\le 0
\right]
=
\sum_{\ell\in\{+1,-1\}}
\Pr\!\left[
L\,\bm a^\top\bm D\le 0
\mid L=\ell
\right]
\Pr[L=\ell].
\]
Combining this identity with \eqref{eq:app_plus_bound} and \eqref{eq:app_minus_bound}, we obtain
\[
\Pr\!\left[
L\,\bm a^\top\bm D\le 0
\right]
\le
\max_{\ell\in\{+1,-1\}}
\exp\!\left(
-\frac{(\bm a^\top\bm\eta)^2}
{2\bm a^\top\bm\Sigma_{\ell}\bm a}
\right).
\]
Since the exponential is monotone and the numerator does not depend on $\ell$, this becomes
\[
\Pr\!\left[
L\,\bm a^\top\bm D\le 0
\right]
\le
\exp\!\left(
-\frac{(\bm a^\top\bm\eta)^2}
{2\max_{\ell\in\{+1,-1\}}\bm a^\top\bm\Sigma_{\ell}\bm a}
\right).
\]
Returning to the original $t$-indexed notation gives
\[
\Pr\!\left[
L_t\,\bm a^\top\bm D_t\le 0
\right]
\le
\exp\!\left(
-\frac{(\bm a^\top\bm\eta_t)^2}
{2\max_{\ell\in\{+1,-1\}}\bm a^\top\bm\Sigma_t(\ell)\bm a}
\right),
\]
which proves \eqref{eq:app_main_bound}. The equivalent form \eqref{eq:app_main_bound_alt} follows from the identity
\[
\frac{(\bm a^\top\bm\eta_t)^2}
{\max_{\ell\in\{+1,-1\}}\bm a^\top\bm\Sigma_t(\ell)\bm a}
=
\min_{\ell\in\{+1,-1\}}
\frac{(\bm a^\top\bm\eta_t)^2}
{\bm a^\top\bm\Sigma_t(\ell)\bm a}.
\]
Defining
\[
\Xi_t(\bm a)
=
\frac{(\bm a^\top\bm\eta_t)^2}
{\max_{\ell\in\{+1,-1\}}\bm a^\top\bm\Sigma_t(\ell)\bm a},
\]
the bound can be written as
\[
\Pr\!\left[
L_t\,\bm a^\top\bm D_t\le 0
\right]
\le
\exp\!\left(-\frac12\Xi_t(\bm a)\right).
\]
Thus, reducing this exponential upper bound is equivalent to maximizing the local surrogate $\Xi_t(\bm a)$, which is the surrogate used in the main text. \hfill $\square$

This motivates the next step: characterizing the scalarizer that maximizes $\Xi_t(\bm a)$ under additional local simplifications.

\subsection{Proof of Theorem~\ref{thm:static_gap_direction}: Static gap and headroom-aware direction}
\label{app:th2_proof}

We now prove the closed-form headroom-aware preferred direction and the resulting local surrogate gap for static aggregation. 
This proof corresponds to Theorem~\ref{thm:static_gap_direction} in the main text.

\paragraph{Restatement of Theorem~\ref{thm:static_gap_direction}.}
Given the local surrogate $\Xi_t(\bm a)$ in Theorem~\ref{thm:local_misallocation}, let $\bm a_t^\star$ denote the locally preferred scalarizer that maximizes $\Xi_t(\bm a)$. 
Under the symmetric and isotropic simplifications and the headroom-shaped edge model in Appendix~\ref{app:pairwise_model}, this preferred scalarizer has the closed-form direction
\begin{equation}
\label{eq:app_headroom_direction}
[\bm a_t^\star]^{(k)}
\propto
w_{\mathrm{base}}^{(k)}
\bigl(H_t^{(k)}\bigr)^{\gamma_0},
\qquad k=1,\ldots,K.
\end{equation}
Relative to this direction, the fixed base scalarizer $\bm w_{\mathrm{base}}$ incurs the local surrogate gap
\begin{equation}
\label{eq:app_static_gap}
\Xi_t(\bm a_t^\star)-\Xi_t(\bm w_{\mathrm{base}})
=
\frac{c_t^2\|\bm w_{\mathrm{base}}\|_2^2}{\sigma_t^2}
\,\mathrm{Var}_{\mu}\!\left(H_t^{\gamma_0}\right),
\end{equation}
where
\begin{equation}
\label{eq:app_mu_def}
\mu^{(k)}
=
\frac{\bigl(w_{\mathrm{base}}^{(k)}\bigr)^2}
{\|\bm w_{\mathrm{base}}\|_2^2},
\qquad
\mathrm{Var}_{\mu}(\bm z)
:=
\sum_{k=1}^{K}\mu^{(k)}(z^{(k)})^2
-
\left(
\sum_{k=1}^{K}\mu^{(k)}z^{(k)}
\right)^2.
\end{equation}
Here $H_t^{\gamma_0}$ denotes the vector whose $k$-th entry is $\bigl(H_t^{(k)}\bigr)^{\gamma_0}$.

\paragraph{Proof.}
The proof has two parts. First, we characterize the locally preferred direction induced by $\Xi_t(\bm a)$. Second, we compare this direction with the fixed base scalarizer $\bm w_{\mathrm{base}}$.

\paragraph{Part I: preferred local direction.}
Under the symmetric residual simplification,
\[
\bm\Sigma_t(+1)=\bm\Sigma_t(-1)=\bm\Sigma_t\succ0,
\]
the local surrogate becomes
\begin{equation}
\label{eq:app_symmetric_surrogate}
\Xi_t(\bm a)
=
\frac{(\bm a^\top\bm\eta_t)^2}
{\bm a^\top\bm\Sigma_t\bm a}.
\end{equation}
For brevity, write
\[
\bm\eta:=\bm\eta_t,
\qquad
\bm\Sigma:=\bm\Sigma_t.
\]
For any nonzero $\bm a\in\mathbb R^K$, define
\[
\bm b:=\bm\Sigma^{1/2}\bm a.
\]
Then
\[
\bm a^\top\bm\Sigma\bm a=\|\bm b\|_2^2,
\qquad
\bm a^\top\bm\eta
=
\bm b^\top\bm\Sigma^{-1/2}\bm\eta.
\]
Substituting these identities into \eqref{eq:app_symmetric_surrogate} gives
\[
\Xi_t(\bm a)
=
\frac{
\left(
\bm b^\top\bm\Sigma^{-1/2}\bm\eta
\right)^2
}
{\|\bm b\|_2^2}.
\]
By Cauchy--Schwarz,
\[
\frac{
\left(
\bm b^\top\bm\Sigma^{-1/2}\bm\eta
\right)^2
}
{\|\bm b\|_2^2}
\le
\|\bm\Sigma^{-1/2}\bm\eta\|_2^2
=
\bm\eta^\top\bm\Sigma^{-1}\bm\eta.
\]
Equality holds if and only if $\bm b$ is a nonzero scalar multiple of $\bm\Sigma^{-1/2}\bm\eta$. 
Equivalently,
\[
\bm\Sigma^{1/2}\bm a
=
\lambda\bm\Sigma^{-1/2}\bm\eta
\]
for some $\lambda\neq0$, which implies
\begin{equation}
\label{eq:app_general_preferred_direction}
\bm a
=
\lambda\bm\Sigma^{-1}\bm\eta.
\end{equation}
Thus, under the symmetric simplification, the unconstrained maximizers of $\Xi_t(\bm a)$ over $\bm a\in\mathbb R^K\setminus\{0\}$ are exactly the nonzero scalar multiples of $\bm\Sigma_t^{-1}\bm\eta_t$.

Under the additional isotropic simplification,
\[
\bm\Sigma_t=\sigma_t^2\bm I,
\qquad
\sigma_t^2>0,
\]
we have
\[
\bm\Sigma_t^{-1}\bm\eta_t
=
\frac{1}{\sigma_t^2}\bm\eta_t.
\]
Therefore, the preferred direction is proportional to $\bm\eta_t$:
\begin{equation}
\label{eq:app_isotropic_preferred_direction}
\bm a_t^\star\propto\bm\eta_t.
\end{equation}
Since $\bm\eta_t\in\mathbb R_+^K$ by (A1), this direction also satisfies the nonnegative scalarizer constraint.

Finally, under the headroom-shaped edge model (A2),
\[
\eta_t^{(k)}
=
c_t\,w_{\mathrm{base}}^{(k)}
\bigl(H_t^{(k)}\bigr)^{\gamma_0},
\qquad k=1,\ldots,K.
\]
Since $c_t>0$ is a global state-dependent scale, it does not affect the direction of the maximizer. 
Combining this with \eqref{eq:app_isotropic_preferred_direction} gives
\[
[\bm a_t^\star]^{(k)}
\propto
w_{\mathrm{base}}^{(k)}
\bigl(H_t^{(k)}\bigr)^{\gamma_0},
\qquad k=1,\ldots,K,
\]
which proves the headroom-aware preferred direction in \eqref{eq:app_headroom_direction}.

\paragraph{Part II: static aggregation gap.}
We now compare the preferred direction $\bm a_t^\star$ with the fixed base scalarizer $\bm w_{\mathrm{base}}$. 
Let
\[
\bm w:=\bm w_{\mathrm{base}},
\qquad
Z^{(k)}:=\bigl(H_t^{(k)}\bigr)^{\gamma_0},
\qquad
\bm Z:=\bigl(Z^{(1)},\ldots,Z^{(K)}\bigr).
\]
Under (A2),
\[
\bm\eta_t
=
c_t(\bm w\odot\bm Z),
\]
where $\odot$ denotes elementwise multiplication.

Since $\bm\Sigma_t=\sigma_t^2\bm I$, the surrogate becomes
\[
\Xi_t(\bm a)
=
\frac{(\bm a^\top\bm\eta_t)^2}
{\sigma_t^2\|\bm a\|_2^2}.
\]
For the preferred direction, Part I gives $\bm a_t^\star\propto\bm\eta_t$. 
Therefore,
\begin{equation}
\label{eq:app_preferred_value}
\Xi_t(\bm a_t^\star)
=
\frac{\|\bm\eta_t\|_2^2}{\sigma_t^2}.
\end{equation}
For the fixed base scalarizer $\bm w$, we have
\begin{equation}
\label{eq:app_static_value}
\Xi_t(\bm w)
=
\frac{(\bm w^\top\bm\eta_t)^2}
{\sigma_t^2\|\bm w\|_2^2}.
\end{equation}
Using $\bm\eta_t=c_t(\bm w\odot\bm Z)$, we compute
\[
\|\bm\eta_t\|_2^2
=
c_t^2
\sum_{k=1}^{K}
\bigl(w_{\mathrm{base}}^{(k)}\bigr)^2
\bigl(Z^{(k)}\bigr)^2,
\]
and
\[
\bm w^\top\bm\eta_t
=
c_t
\sum_{k=1}^{K}
\bigl(w_{\mathrm{base}}^{(k)}\bigr)^2
Z^{(k)}.
\]
Substituting these expressions into \eqref{eq:app_preferred_value} and \eqref{eq:app_static_value} yields
\[
\Xi_t(\bm a_t^\star)-\Xi_t(\bm w)
=
\frac{c_t^2}{\sigma_t^2}
\left[
\sum_{k=1}^{K}
\bigl(w_{\mathrm{base}}^{(k)}\bigr)^2
\bigl(Z^{(k)}\bigr)^2
-
\frac{
\left(
\sum_{k=1}^{K}
\bigl(w_{\mathrm{base}}^{(k)}\bigr)^2
Z^{(k)}
\right)^2
}
{\|\bm w_{\mathrm{base}}\|_2^2}
\right].
\]
Now define the base-induced weighting distribution
\[
\mu^{(k)}
=
\frac{\bigl(w_{\mathrm{base}}^{(k)}\bigr)^2}
{\|\bm w_{\mathrm{base}}\|_2^2},
\qquad k=1,\ldots,K.
\]
Since $\sum_{k=1}^{K}\mu^{(k)}=1$, the previous expression becomes
\[
\Xi_t(\bm a_t^\star)-\Xi_t(\bm w)
=
\frac{c_t^2\|\bm w_{\mathrm{base}}\|_2^2}{\sigma_t^2}
\left[
\sum_{k=1}^{K}\mu^{(k)}\bigl(Z^{(k)}\bigr)^2
-
\left(
\sum_{k=1}^{K}\mu^{(k)}Z^{(k)}
\right)^2
\right].
\]
By the definition of weighted variance under $\mu$,
\[
\mathrm{Var}_{\mu}(\bm Z)
=
\sum_{k=1}^{K}\mu^{(k)}\bigl(Z^{(k)}\bigr)^2
-
\left(
\sum_{k=1}^{K}\mu^{(k)}Z^{(k)}
\right)^2,
\]
we obtain
\[
\Xi_t(\bm a_t^\star)-\Xi_t(\bm w_{\mathrm{base}})
=
\frac{c_t^2\|\bm w_{\mathrm{base}}\|_2^2}{\sigma_t^2}
\,\mathrm{Var}_{\mu}(\bm Z).
\]
Recalling that $Z^{(k)}=\bigl(H_t^{(k)}\bigr)^{\gamma_0}$, this is exactly
\[
\Xi_t(\bm a_t^\star)-\Xi_t(\bm w_{\mathrm{base}})
=
\frac{c_t^2\|\bm w_{\mathrm{base}}\|_2^2}{\sigma_t^2}
\,\mathrm{Var}_{\mu}\!\left(H_t^{\gamma_0}\right),
\]
which proves the static aggregation gap in \eqref{eq:app_static_gap}. \hfill $\square$

\paragraph{Remark.}
The variance term is zero if the headroom factor $\bigl(H_t^{(k)}\bigr)^{\gamma_0}$ is effectively constant over the criteria that receive nonzero base weight. 
In that case, the preferred direction is only a scalar multiple of $\bm w_{\mathrm{base}}$, so static aggregation remains locally aligned with the preferred direction. 
When the base-weighted headroom profile is heterogeneous, the variance term becomes positive, and the fixed base scalarizer incurs a positive local surrogate gap relative to the headroom-aware direction.

\subsection{Bridge to the implementation proxy}
\label{app:proxy_bridge}

The main text implements the headroom-aware direction using the saturation
statistic \(P^{(k)}\) defined in Section~\ref{sec:focal_reward}. We restate the
same quantities here with an explicit training-state timestep index \(t\), which will be
used in the proof below.

At training state \(t\), consider a sampled rollout group
\(\{y_i\}_{i=1}^G\). Let
\[
R^{\mathrm{base}}_{t,i}
:=
R_i(\bm w_{\mathrm{base}})
\]
denote the scalar reward of rollout \(y_i\) computed by the base scalarizer.
The corresponding Gibbs rollout weight is
\[
\widehat r_{t,i}
:=
\frac{
\exp(R^{\mathrm{base}}_{t,i}/T)
}{
\sum_{n=1}^G \exp(R^{\mathrm{base}}_{t,n}/T)
},
\qquad
\sum_{i=1}^G\widehat r_{t,i}=1 .
\]
For criterion \(c^{(k)}\), define
\[
\bar s_{t,i}^{(k)}
:=
\frac{1}{G-1}\sum_{j\ne i}s_{i,j}^{(k)},
\qquad
z_{t,i}^{(k)}
:=
\frac{\bar s_{t,i}^{(k)}}{S_{\max}} .
\]
Since \(s_{i,j}^{(k)}\in[0,S_{\max}]\), we have
\(z_{t,i}^{(k)}\in[0,1]\). The time-indexed form of the implemented saturation
statistic is
\[
P_t^{(k)}
:=
\sum_{i=1}^G \widehat r_{t,i}z_{t,i}^{(k)}
=
\frac{
\langle \widehat{\bm r}_t,\bar{\bm s}_t^{(k)}\rangle
}{
S_{\max}
}.
\]
Thus \(P_t^{(k)}\) is exactly the inverse reward projection statistic used in
the main algorithm. It measures the criterion saturation level under the
base-reward-weighted rollout distribution, and \(1-P_t^{(k)}\) is used as the
observable proxy for remaining frontier headroom.

The relevant quantity in Theorem~\ref{thm:static_gap_direction} is frontier headroom, not
the average dissatisfaction over all sampled rollouts. A global average over the
entire rollout group can be dominated by low-quality responses and therefore
reflect broad response weakness rather than the residual bottleneck of the
current strong responses. In contrast, policy improvement is driven by how the
scalar reward assigns credit among competitive rollouts. Therefore, the
saturation statistic should emphasize criteria that remain under-satisfied on
the current reward frontier. The Gibbs weighting used in the implementation
provides a soft way to estimate this frontier-level saturation without
introducing a hard top-\(k\) selection rule.

\begin{proposition}[Gibbs frontier concentration]
\label{prop:app_gibbs_bridge}
Let \(T>0\) and fix a nonempty proper subset
\(\mathcal S_t\subset\{1,\ldots,G\}\), with
\(1\le |\mathcal S_t|<G\). Suppose there exists a reward gap
\(\delta_t>0\) such that
\begin{equation}
\label{eq:app_frontier_gap}
\min_{i\in \mathcal S_t} R^{\mathrm{base}}_{t,i}
\ge
\max_{j\notin \mathcal S_t} R^{\mathrm{base}}_{t,j}
+
\delta_t .
\end{equation}
Let
\[
\rho_t:=\sum_{i\in\mathcal S_t}\widehat r_{t,i}
\]
denote the Gibbs mass assigned to \(\mathcal S_t\). Then
\begin{equation}
\label{eq:app_frontier_mass}
1-\rho_t
\le
\frac{G-|\mathcal S_t|}{|\mathcal S_t|}
\exp(-\delta_t/T).
\end{equation}
Moreover, for every criterion \(c^{(k)}\), define the
frontier-restricted saturation estimate
\[
P_{t,\mathcal S}^{(k)}
:=
\sum_{i\in\mathcal S_t}
\frac{\widehat r_{t,i}}{\sum_{m\in\mathcal S_t}\widehat r_{t,m}}
z_{t,i}^{(k)} .
\]
Then
\begin{equation}
\label{eq:app_frontier_proxy}
\left|
P_t^{(k)}-P_{t,\mathcal S}^{(k)}
\right|
\le
\frac{G-|\mathcal S_t|}{|\mathcal S_t|}
\exp(-\delta_t/T).
\end{equation}
\end{proposition}

\paragraph{Proof.}
Define
\[
\alpha
:=
\sum_{i\in\mathcal S_t}
\exp(R^{\mathrm{base}}_{t,i}/T),
\qquad
\beta
:=
\sum_{j\notin\mathcal S_t}
\exp(R^{\mathrm{base}}_{t,j}/T).
\]
Then
\[
\rho_t
=
\frac{\alpha}{\alpha+\beta}
=
\sum_{i\in\mathcal S_t}\widehat r_{t,i}.
\]
By \eqref{eq:app_frontier_gap}, for every \(j\notin\mathcal S_t\),
\[
R^{\mathrm{base}}_{t,j}
\le
\min_{i\in\mathcal S_t}R^{\mathrm{base}}_{t,i}
-
\delta_t .
\]
Therefore,
\[
\exp(R^{\mathrm{base}}_{t,j}/T)
\le
\exp(-\delta_t/T)
\exp\!\left(
\frac{\min_{i\in\mathcal S_t}R^{\mathrm{base}}_{t,i}}{T}
\right).
\]
Summing over \(j\notin\mathcal S_t\) gives
\[
\beta
\le
(G-|\mathcal S_t|)
\exp(-\delta_t/T)
\exp\!\left(
\frac{\min_{i\in\mathcal S_t}R^{\mathrm{base}}_{t,i}}{T}
\right).
\]
Also,
\[
\alpha
\ge
|\mathcal S_t|
\exp\!\left(
\frac{\min_{i\in\mathcal S_t}R^{\mathrm{base}}_{t,i}}{T}
\right).
\]
Hence
\[
\frac{\beta}{\alpha}
\le
\frac{G-|\mathcal S_t|}{|\mathcal S_t|}
\exp(-\delta_t/T).
\]
Since
\[
1-\rho_t
=
\frac{\beta}{\alpha+\beta}
\le
\frac{\beta}{\alpha},
\]
we obtain \eqref{eq:app_frontier_mass}.

Next, define the out-of-frontier saturation estimate
\[
P_{t,\mathrm{out}}^{(k)}
:=
\frac{1}{1-\rho_t}
\sum_{j\notin\mathcal S_t}
\widehat r_{t,j}z_{t,j}^{(k)}
\]
when \(1-\rho_t>0\). If \(1-\rho_t=0\), choose any
\(P_{t,\mathrm{out}}^{(k)}\in[0,1]\). Since every
\(z_{t,i}^{(k)}\in[0,1]\), both \(P_{t,\mathcal S}^{(k)}\) and
\(P_{t,\mathrm{out}}^{(k)}\) lie in \([0,1]\). Moreover,
\[
P_t^{(k)}
=
\sum_{i\in\mathcal S_t}\widehat r_{t,i}z_{t,i}^{(k)}
+
\sum_{j\notin\mathcal S_t}\widehat r_{t,j}z_{t,j}^{(k)}
=
\rho_t P_{t,\mathcal S}^{(k)}
+
(1-\rho_t)P_{t,\mathrm{out}}^{(k)} .
\]
Therefore,
\[
P_t^{(k)}-P_{t,\mathcal S}^{(k)}
=
(1-\rho_t)
\left(
P_{t,\mathrm{out}}^{(k)}-P_{t,\mathcal S}^{(k)}
\right).
\]
Taking absolute values gives
\[
\left|
P_t^{(k)}-P_{t,\mathcal S}^{(k)}
\right|
\le
1-\rho_t .
\]
Combining this with \eqref{eq:app_frontier_mass} yields
\eqref{eq:app_frontier_proxy}. \(\square\)

\paragraph{Why not use a global average?}
If one instead used the unweighted global statistic
\[
P_{t,\mathrm{global}}^{(k)}
=
\frac{1}{G}\sum_{i=1}^G z_{t,i}^{(k)},
\]
then its deviation from a frontier-restricted average would generally scale with
the fraction of non-frontier rollouts rather than with the reward gap. For
example, writing
\[
P_{t,\mathrm{global}}^{(k)}
=
\frac{|\mathcal S_t|}{G}P_{t,\mathcal S,\mathrm{unif}}^{(k)}
+
\left(1-\frac{|\mathcal S_t|}{G}\right)
P_{t,\mathrm{out},\mathrm{unif}}^{(k)},
\]
the coefficient on the out-of-frontier average is
\(1-|\mathcal S_t|/G\), which can be large even when the base reward clearly
separates strong and weak rollouts. By contrast, Proposition~\ref{prop:app_gibbs_bridge}
shows that the out-of-frontier contribution under Gibbs weighting is bounded by
\[
\frac{G-|\mathcal S_t|}{|\mathcal S_t|}\exp(-\delta_t/T),
\]
which decreases as the frontier gap grows. Thus the Gibbs statistic estimates
frontier saturation, while the global average estimates overall rollout-pool
saturation.

\paragraph{Interpretation.}
Proposition~\ref{prop:app_gibbs_bridge} shows that when the base-pass rewards
exhibit a nontrivial frontier gap, the Gibbs weights \(\widehat r_{t,i}\)
concentrate on the strong-rollout subset \(\mathcal S_t\). In this regime,
\(P_t^{(k)}\) is close to the criterion saturation level computed only on the
reward frontier. Therefore, \(1-P_t^{(k)}\) can be interpreted as an operational
proxy for the remaining frontier headroom of criterion \(c^{(k)}\), which is
the observable counterpart of the latent headroom \(H_t^{(k)}\) used in
Theorem~\ref{thm:static_gap_direction}.

\subsection{Identification boundary of the implementation statistic}
\label{app:nonidentifiability}

The previous analysis justifies the headroom-aware scalarizer direction through
criterion-level pairwise differences. However, estimating saturation requires
absolute criterion-score levels from the full judge score tensor. The following
proposition clarifies this boundary.

\begin{proposition}[Pairwise differences do not identify \(P_t^{(k)}\)]
\label{prop:app_nonidentifiability}
Fix a criterion \(c^{(k)}\). Suppose there exists a constant \(b_k\neq 0\) such
that the shifted scores
\[
s_{i,j}^{\prime(k)}:=s_{i,j}^{(k)}+b_k
\]
remain in \([0,S_{\max}]\) for all ordered pairs \((i,j)\) with \(i\neq j\).
For every \(k'\neq k\), let
\[
s_{i,j}^{\prime(k')}:=s_{i,j}^{(k')}.
\]
Then all pairwise difference vectors, base margins, base rewards, and Gibbs
rollout weights remain unchanged:
\[
\bm d'_{i,j}=\bm d_{i,j},
\qquad
\Delta'_{i,j}(\bm w_{\mathrm{base}})
=
\Delta_{i,j}(\bm w_{\mathrm{base}}),
\qquad
R_{t,i}^{\mathrm{base}\prime}
=
R_{t,i}^{\mathrm{base}},
\qquad
\widehat r'_{t,i}
=
\widehat r_{t,i}.
\]
However, the saturation statistic for criterion \(c^{(k)}\) changes as
\[
P_t^{\prime(k)}
=
P_t^{(k)}+\frac{b_k}{S_{\max}}.
\]
Therefore, \(P_t^{(k)}\) is not identified by the pairwise difference vectors
alone.
\end{proposition}

\paragraph{Proof.}
For criterion \(c^{(k)}\),
\[
d_{i,j}^{\prime(k)}
=
s_{i,j}^{\prime(k)}-s_{j,i}^{\prime(k)}
=
(s_{i,j}^{(k)}+b_k)-(s_{j,i}^{(k)}+b_k)
=
d_{i,j}^{(k)}.
\]
For every \(k'\neq k\), the scores are unchanged, so
\[
d_{i,j}^{\prime(k')}=d_{i,j}^{(k')}.
\]
Hence \(\bm d'_{i,j}=\bm d_{i,j}\) for all ordered pairs. Since the base margin
depends on the score tensor only through \(\bm d_{i,j}\), the base margins are
unchanged. The mapped pairwise outcomes and base rewards are therefore
unchanged, and the Gibbs rollout weights, as deterministic functions of the base
rewards, are unchanged as well.

On the other hand,
\[
\bar s_{t,i}^{\prime(k)}
=
\frac{1}{G-1}\sum_{j\neq i}s_{i,j}^{\prime(k)}
=
\bar s_{t,i}^{(k)}+b_k .
\]
Thus
\[
P_t^{\prime(k)}
=
\frac{1}{S_{\max}}
\sum_{i=1}^G
\widehat r'_{t,i}\bar s_{t,i}^{\prime(k)}
=
\frac{1}{S_{\max}}
\sum_{i=1}^G
\widehat r_{t,i}(\bar s_{t,i}^{(k)}+b_k)
=
P_t^{(k)}+\frac{b_k}{S_{\max}},
\]
where we use \(\sum_{i=1}^G\widehat r_{t,i}=1\). This proves the claim.
\(\square\)

\paragraph{Remark.}
Proposition~\ref{prop:app_nonidentifiability} only clarifies the scope of the
theoretical analysis. The pairwise difference model supports the local
headroom-aware scalarizer direction, while the implemented statistic
\(P_t^{(k)}\) uses additional absolute score-level information from the full
judge score tensor. Thus \(P_t^{(k)}\) should be viewed as an operational
saturation proxy with the frontier interpretation established in
Proposition~\ref{prop:app_gibbs_bridge}, rather than as a latent headroom value
identified from pairwise differences alone.

\section{Detailed Training Settings}
\label{app:training_details}

For RL post-training, we train on the corresponding OpenRubrics split using the verl framework~\citep{DBLP:conf/eurosys/ShengZYWZZPL025}. To preserve domain-specific characteristics, training is performed separately for each setting for 3 epochs with GSPO. We use a batch size of 64 (mini-batch 32) and AdamW with a fixed learning rate of $1\times10^{-6}$ and weight decay $0.1$. A KL divergence penalty (low-variance estimator) with coefficient $0.001$ is added to the actor loss to prevent excessive deviation from the reference policy. For each prompt, 8 rollouts are sampled with temperature $1.0$ and top-$p$ $1.0$. The maximum prompt and response lengths are 8{,}192 and 12{,}000 tokens, respectively. Clipping bounds are set to $\epsilon_{\mathrm{low}} = 3\times10^{-4}$ and $\epsilon_{\mathrm{high}} = 4\times10^{-4}$. Key hyper-parameters are summarized in Table~\ref{tab:rl_training_config}.

\paragraph{Compute resources.}
All experiments are conducted on H800-80GB GPUs. Qwen2.5-7B-Instruct experiments use 8 GPUs, Qwen3-8B experiments use 16 GPUs, and Qwen3-30B-A3B experiments use 32 GPUs. Each training run uses 500 total training steps, and representative end-to-end step times and reward-synthesis overhead are reported in Table~\ref{tab:runtime}.

\begin{table}[t]
\centering
\small
\caption{RL training configuration.}
\label{tab:rl_training_config}
\begin{tabular}{ll}
\toprule
Category & Configuration \\
\midrule
\textbf{GSPO} & RL Algorithm: GSPO \\
& Clip $\epsilon_{\mathrm{low}}$ / $\epsilon_{\mathrm{high}}$: $3\times10^{-4}$ / $4\times10^{-4}$ \\
\midrule
\textbf{Backbone Model} & Policy Models: Qwen2.5-7B-Instruct, Qwen3-8B, Qwen3-30B-A3B \\
& Judge Model: Qwen3-80B-A3B-Instruct \\
\midrule
\textbf{Sampling} & Train Temperature: 1.0 \\
& Train Top-P: 1.0 \\
& Rollout Samples per Prompt ($G$): 8 \\
& Max Prompt Length: 8{,}192 \\
& Max Response Length: 12{,}000 \\
\midrule
\textbf{Training} & Optimizer: AdamW \\
& Learning Rate: $1\times10^{-6}$ \\
& Weight Decay: 0.1 \\
& Training Batch Size: 64 \\
& Mini Batch Size: 32 \\
& KL Loss Coefficient: 0.001 \\
& KL Loss Type: Low-Variance KL \\
& Total Training Steps: 500 \\
& Total Epochs: 3 \\
& $\tau$ for $\Phi_\tau(\Delta)$ in Eq~\ref{eq:group_reward}: 1 \\
\midrule
\textbf{Hardware} & GPUs for Qwen2.5-7B-Instruct: 8 $\times$ H800-80GB \\
& GPUs for Qwen3-8B: 16 $\times$ H800-80GB \\
& GPUs for Qwen3-30B-A3B: 32 $\times$ H800-80GB \\
\bottomrule
\end{tabular}
\end{table}

\FloatBarrier

\paragraph{Evaluation benchmarks.}
We evaluate general-domain models on four complementary open-ended benchmarks. 
AlpacaEval 2.0 measures instruction-following quality through automatic pairwise evaluation and reports length-controlled win rates, which helps reduce the known length bias of automatic evaluators~\citep{DBLP:journals/corr/abs-2404-04475}. 
Arena-Hard evaluates challenging open-ended prompts curated from real user interactions and is designed to provide a stronger separation among modern chat models~\citep{DBLP:conf/icml/LiCFD0ZGS25}. 
WritingBench focuses on generative writing, covering diverse writing requirements across multiple writing domains and subdomains; we include it because rubric-based reward synthesis is especially relevant for long-form and style-sensitive generation~\citep{DBLP:journals/corr/abs-2503-05244}. 
EQ-Bench 3 evaluates emotional intelligence, social reasoning, empathy, and interpersonal response quality in multi-turn or role-play settings; it complements generic instruction-following benchmarks by testing socially nuanced open-ended behavior~\citep{DBLP:journals/corr/abs-2312-06281}.

For science-oriented evaluation, we use GPQA Diamond and HealthBench. 
GPQA Diamond is a difficult subset of GPQA, a graduate-level, Google-proof multiple-choice benchmark written by domain experts in biology, physics, and chemistry; it tests expert-level scientific reasoning rather than general chat quality~\citep{rein2024gpqa}. 
HealthBench evaluates model behavior in healthcare conversations using conversation-specific rubrics authored by physicians, covering both performance and safety in realistic health-related interactions~\citep{DBLP:journals/corr/abs-2505-08775}. 
Together, these benchmarks cover instruction following, hard open-ended preference evaluation, writing quality, social-emotional reasoning, expert scientific reasoning, and safety-critical healthcare dialogue.

\paragraph{Evaluation repetitions.}
For LLM-based evaluation benchmarks, we repeat each evaluation multiple times and report the averaged score to reduce judge stochasticity. This repeated-evaluation protocol is used for the main benchmark results reported in Table~\ref{tab:main_results} and Table~\ref{tab:ablation}.

\section{Prompt Templates}
\label{app:prompt_templates}
This section documents the two main prompt templates used in our system: the pairwise judge prompt (Section~\ref{app:judge_prompt}) and the prior-weight generation prompt (Section~\ref{app:weight_prompt}).

\subsection{Pairwise Judge Prompt}
\label{app:judge_prompt}

For each ordered pair of rollouts $(y_i, y_j)$ within a rollout group, we query the judge model (Qwen3-80B-A3B-Instruct) to obtain criterion-level scores. The prompt instructs the judge to evaluate both rollouts against the full rubric and return per-criterion scores in a structured JSON format. Criteria are divided into two types with distinct scoring semantics:
\begin{itemize}[nosep,leftmargin=*]
    \item \textbf{[Hard Rule]} criteria enforce binary compliance (factuality, safety, format constraints). The judge outputs \texttt{true} (compliant, mapped to $10$) or \texttt{false} (violation, mapped to $0$).
    \item \textbf{[Principle]} criteria assess soft quality dimensions (logical depth, coherence, helpfulness). The judge outputs an integer score from $0$ to $10$.
\end{itemize}
To mitigate position bias, we evaluate each pair in both orders and average the resulting scores. The template below is instantiated with the user question, the two rollouts, and the prompt-specific rubric.

\begin{tcblisting}{
  colback=gray!3,
  colframe=gray!50,
  coltitle=white,
  colbacktitle=gray!50,
  fonttitle=\bfseries,
  title=Pairwise Judge Prompt,
  listing only,
  breakable,
  enhanced,
  listing options={
    basicstyle=\ttfamily\small,
    columns=fullflexible,
    keepspaces=true,
    breaklines=true,
    breakatwhitespace=true,
    showstringspaces=false
  }
}
# Role and Objective
You are an expert evaluator. Your task is to evaluate two AI assistant rollouts, Rollout A and Rollout B, according to the given Evaluation rubric. Score both rollouts item by item using the rubric order.

# Input Data
## User Question
<|begin_of_query|>
{question}
<|end_of_query|>

## Rollout A
<|begin_of_rollout_A|>
{prediction1}
<|end_of_rollout_A|>

## Rollout B
<|begin_of_rollout_B|>
{prediction2}
<|end_of_rollout_B|>

# Evaluation rubric
The rubric contains {count} criteria. Each criterion is marked as either [Hard Rule] or [Principle].

<|begin_of_rubric|>
{rubric}
<|end_of_rubric|>

# Scoring Rules
For each criterion, evaluate Rollout A and Rollout B independently.

## [Hard Rule]
A [Hard Rule] criterion checks binary compliance with requirements such as factuality, safety, format constraints, or explicit user instructions.

Output only true or false:
1. true means the rollout fully satisfies the criterion.
2. false means the rollout violates the criterion.

Do not output numeric scores for [Hard Rule] criteria.

## [Principle]
A [Principle] criterion evaluates graded rollout quality, such as helpfulness, completeness, reasoning quality, clarity, structure, or depth.

Output an integer score from 0 to 10:
1. 9 to 10 means outstanding.
2. 6 to 8 means good.
3. 3 to 5 means mediocre.
4. 0 to 2 means poor.

# Evaluation Procedure
1. Check whether each rollout violates any [Hard Rule] criteria.
2. Evaluate both rollouts on each [Principle] criterion.
3. Produce a brief item level rationale.
4. Convert the judgments into the required true, false, or numeric scores.

# Output Format
Return ONLY a JSON object in the following format:
{
  "rationale": "Brief item level comparison of Rollout A and Rollout B according to the rubric.",
  "rollout_A_scores": [true, 8, false, 5],
  "rollout_B_scores": [true, 6, true, 7]
}

Requirements:
1. The length of each score list must equal {count}.
2. Scores must follow the same order as the rubric.
3. [Hard Rule] positions must be boolean.
4. [Principle] positions must be integers from 0 to 10.
\end{tcblisting}

\subsection{Prior-Weight Generation Prompt}
\label{app:weight_prompt}

For the \emph{static\_prior\_weight} and \emph{focal\_prior\_weight} reward modes, we use GPT-4o (\texttt{gpt-4o-2024-11-20}) to generate question-specific base weights $\bm w_{\mathrm{base}}$ before training. Given a user question and its associated rubric, the model is asked to assign an importance weight to each criterion reflecting its relevance to the specific question. The returned weights are normalized to sum to one and stored alongside the training data. This procedure is run once per training example as a preprocessing step; the weights remain fixed throughout training for the static modes, and serve as the base weights that Focal Reward dynamically modulates .

\begin{tcblisting}{
  colback=gray!3,
  colframe=gray!50,
  coltitle=white,
  colbacktitle=gray!50,
  fonttitle=\bfseries,
  title=Prior Weight Generation Prompt (GPT-4o),
  listing only,
  breakable,
  enhanced,
  listing options={
    basicstyle=\ttfamily\small,
    columns=fullflexible,
    keepspaces=true,
    breaklines=true,
    breakatwhitespace=true,
    showstringspaces=false
  }
}
You are an expert evaluator. Your task is to assign importance weights to the evaluation criteria for a given user question.

## User Question
{prompt}

## Evaluation Criteria
{rubric_str}

## Criterion Types
The criteria may include two types:
1. [Hard Rule]: binary compliance criteria, such as factuality, safety, format constraints, or explicit requirements.
2. [Principle]: graded quality criteria, such as helpfulness, completeness, reasoning quality, clarity, or depth.

## Instructions
1. Assign one weight to each criterion according to how relevant and important it is for evaluating the answer to this specific user question.
2. Consider both the content of the user question and the role of each criterion in the rubric.
3. The weights should be nonnegative.
4. The sum of all weights should equal 1.0.

## Output Format
Return ONLY a JSON object with no explanation:
{"weights": [w1, w2, w3, ...]}

Each weight must correspond to the criterion at the same position in the rubric above.
\end{tcblisting}

\section{Case Study}
\label{app:case_study}

To provide qualitative evidence that the statistical patterns from Section~\ref{sec:main_results} manifest in concrete examples, we select two prompts from the Qwen2.5-7B-Instruct rollouts where the differences between static and focal training are directly visible in the rollout text. Below, for each case we show the criterion-level mechanism (epoch-1 $P^{(k)}$) alongside the outcome (epoch-3 best-rollout scores under each run), followed by excerpted rollouts.

\subsection{Case 1: Instruction following---first-person persona}

\paragraph{Prompt.} \textit{``Can you pretend a man called `Tong Chi Ming' and living in UK?''}

The user's intent is clear: ``pretend a man'' requests the model to \emph{adopt a first-person persona}, not to create a third-person character profile.  This prompt has 6 criteria: 1 hard rule and 5 principles.  The hard rule---\emph{written in first person as a man named Tong Chi Ming residing in the UK}---has the lowest saturation ($P^{(k)} = 0.55$), while the remaining criteria (transparency, detail richness, avoiding fabrication, substantive content, minimal refusal) are near-saturated ($P^{(k)} = 0.80$--$0.92$).  Table~\ref{tab:case_10213} shows the full criterion profile.

\begin{table}[h]
\centering
\small
\setlength{\tabcolsep}{3pt}
\caption{Case 1 (instruction following): criterion-level mechanism and outcome. Bold indicates the under-saturated hard criterion.}
\label{tab:case_10213}
\begin{tabular}{p{0.50\textwidth}ccccc}
\toprule
Criterion & Type & $P^{(k)}$ & Static & Focal & $\Delta$ \\
\midrule
\textbf{Written in first person as Tong Chi Ming in UK} & H & \textbf{0.55} & 2.0 & \textbf{10.0} & \textbf{+8.0} \\
Transparently clarify capabilities/limitations & P & 0.80 & 7.9 & 7.7 & $-$0.2 \\
Contextually relevant persona details & P & 0.82 & 7.8 & 8.6 & $+$0.8 \\
Avoid fabricating mistakable personal info & P & 0.83 & 8.6 & 8.6 & $+$0.0 \\
Substantive, coherent content & P & 0.88 & 8.6 & 8.6 & $+$0.0 \\
Refusal only when necessary & P & 0.92 & 9.0 & 8.8 & $-$0.2 \\
\bottomrule
\end{tabular}
\end{table}

The critical observation is that both rollouts are substantive and address the user's request, yet they adopt fundamentally different rollout \emph{frameworks}.  The static-trained rollout opens with ``Let's create a profile for Tong Chi Ming, a fictional person'' and proceeds entirely in the third person (``He moved\ldots'', ``His expertise\ldots''), scoring 2.0 on the hard criterion.  The focal-trained rollout opens with ``Hello, my name is Tong Chi Ming'' and maintains a first-person persona throughout (``I moved\ldots'', ``I work\ldots''), scoring 10.0.  This is a $+8.0$ point improvement on the core instruction---the maximum possible gain---with no regression on any other criterion (mean easy $\Delta = -0.05$).  Notably, the focal rollout is also 35\% shorter (1{,}924 vs.\ 2{,}958 characters), demonstrating that the improvement comes from correctly understanding the instruction, not from generating more text.

\paragraph{Rollout outlines.}
The core difference is visible in the first sentence of each rollout.

\smallskip
\noindent
\begin{minipage}[t]{0.48\textwidth}
\begin{tcolorbox}[colback=gray!5, colframe=gray!50, fontupper=\footnotesize, title={\footnotesize Static (2{,}958 chars)}, fonttitle=\bfseries\footnotesize, boxrule=0.3pt, left=2pt, right=2pt, top=1pt, bottom=1pt]
\textcolor{red!70!black}{\textbf{``Certainly! Let's create a profile for Tong Chi Ming, a fictional person living in the UK.''}}\\[3pt]
$\to$ Third-person character profile\\
$\to$ ``He moved\ldots'', ``His expertise\ldots''\\
$\to$ Structured as a biography/resume\\
$\to$ Ends: ``Feel free to ask for more details \textbf{involving} Tong Chi Ming!''
\end{tcolorbox}
\end{minipage}\hfill
\begin{minipage}[t]{0.48\textwidth}
\begin{tcolorbox}[colback=blue!3, colframe=blue!30, fontupper=\footnotesize, title={\footnotesize Focal (1{,}924 chars)}, fonttitle=\bfseries\footnotesize, boxrule=0.3pt, left=2pt, right=2pt, top=1pt, bottom=1pt]
\textcolor{blue!60!black}{\textbf{``Certainly! Hello, my name is Tong Chi Ming, and I live in the United Kingdom.''}}\\[3pt]
$\to$ First-person persona\\
$\to$ ``I moved\ldots'', ``I work\ldots''\\
$\to$ Natural self-introduction\\
$\to$ Ends: ``feel free to reach out!''
\end{tcolorbox}
\end{minipage}

\smallskip\noindent
The static rollout interprets ``pretend a man'' as ``create a character profile'' (third person), while the focal rollout correctly interprets it as ``adopt a persona'' (first person).  Under base reward, the third-person profile scores well on surface-level criteria---structured formatting, detailed information, explicit fictional disclaimers---and is selected as the best rollout despite completely misunderstanding the core instruction.  Focal reweighting raises the hard criterion's weight from 37.5\% to 78.2\%, causing the first-person rollout to be correctly selected as best.

\paragraph{Static-trained rollout.}  \textcolor{red!70!black}{Highlighted} marks the third-person framing that violates the core instruction.
\begin{tcolorbox}[colback=gray!5, colframe=gray!50, fontupper=\small, breakable, left=4pt, right=4pt, top=3pt, bottom=3pt]
\textcolor{red!70!black}{Certainly! Let's create a profile for Tong Chi Ming, a fictional person living in the UK. Here's a fabricated background for him:}

---

\noindent\textbf{Name:} Tong Chi Ming\\
\textbf{Age:} 34\\
\textbf{Nationality:} Hong Kong Chinese (now settled in the UK)\\
\textbf{Place of Residence:} London, England\\
\textbf{Occupation:} Business Consultant\\
\textbf{Education:}\\
$\bullet$~Bachelor of Science in Business Administration, University of Hong Kong (2008-2012)\\
$\bullet$~Master of Business Administration, London Business School (2013-2014)

\noindent\textbf{Background:}
\textcolor{red!70!black}{Tong Chi Ming \textbf{was} born and raised in Hong Kong but moved to the United Kingdom in 2012 to pursue \textbf{his} Master's degree at the prestigious London Business School.  After graduating, \textbf{he} began \textbf{his} career in consultancy, working for a leading international firm headquartered in London.  \textbf{His} expertise lies in strategic planning, market analysis, and leveraging digital transformation to help businesses achieve their goals.}

\noindent\textbf{Current Career:}
Tong Chi Ming \textbf{is} currently a senior business consultant with a renowned UK-based consulting firm.  \textbf{His} role involves advising both emerging and established companies on strategic decision-making, risk management, and operational efficiency.  \textbf{He} has a particular interest in technology-driven solutions and has spearheaded several projects that have successfully enabled companies to adopt digital ecosystems and innovate their services.

\noindent\textbf{Interests:}\\
$\bullet$~Technology and innovation trends\\
$\bullet$~Travel (especially to Asia and Europe)\\
$\bullet$~Literature and philosophy\\
$\bullet$~Volunteering for community development and environmental conservation projects\\
$\bullet$~Cooking and fine dining

\noindent\textbf{Hobbies:}\\
$\bullet$~Exploring new restaurants and trying out different cuisines around London\\
$\bullet$~Attending cultural and educational events (concerts, lectures, exhibitions)\\
$\bullet$~Going for regular runs in Hyde Park to clear his mind and stay fit\\
$\bullet$~Participating in community service projects, particularly those focused on sustainability and urban green spaces

\noindent\textbf{Family:}\\
$\bullet$~No immediate family members in the UK, though he keeps in close contact with his parents and siblings in Hong Kong.\\
$\bullet$~Living alone but often visits his family during major holidays or special occasions.

\noindent\textbf{Social Media Presence:}
Tong Chi Ming is moderately active on LinkedIn, where he shares insights on business and technology.  He occasionally posts on his personal blog (tongchiming.com) about his experiences in business, travel, and personal development, using a mix of textual and visual content.

\noindent\textbf{Recent Activities:}\\
$\bullet$~Recently completed a project for a London-based tech startup, helping them develop a plan for expanding their international market presence.\\
$\bullet$~Attended a Friedrich Ebert Stiftung (FES) event in London to discuss innovation and entrepreneurship in a post-pandemic world.\\
$\bullet$~Participated in a local running event to raise funds for a charity focused on ocean conservation.

---

Feel free to ask for more details or scenarios involving Tong Chi Ming!
\end{tcolorbox}

\paragraph{Focal-trained rollout.}  \textcolor{blue!60!black}{Highlighted} marks the correct first-person persona framing.
\begin{tcolorbox}[colback=blue!3, colframe=blue!30, fontupper=\small, breakable, left=4pt, right=4pt, top=3pt, bottom=3pt]
\textcolor{blue!60!black}{Certainly! Hello, my name is Tong Chi Ming, and \textbf{I} live in the United Kingdom.  \textbf{I} was born and raised in Hong Kong, but \textbf{I}'ve called the UK \textbf{my} home for the past decade.  \textbf{I} moved to London for university and stayed on after graduation, embracing the vibrant and diverse city that London truly is.}

\textbf{I} work as a software engineer at a tech startup based in London.  \textbf{I} specialize in backend development and enjoy tackling complex problems through code.  \textbf{My} workdays are as varied as they are stimulating, and \textbf{I} feel incredibly fortunate to be able to combine \textbf{my} love for technology with challenging and rewarding projects.

Outside of work, \textbf{I} have a passion for photography.  \textbf{I} love capturing the unique angles and perspectives of the city, and \textbf{I} find it fascinating to see London through the lens of a camera.  On weekends, you might catch \textbf{me} exploring new galleries, sampling local cuisines, or just basking in the quaint pubs with a pint and a good book.

\textbf{I} am also an avid reader, with a particular interest in history and philosophy.  \textbf{I} believe that understanding the past and contemplating broader ideas are essential for personal growth and helping navigate the complexities of the modern world.

When the weather allows, you might find \textbf{me} cycling through the picturesque countryside around London.  \textbf{I}'ve grown to love the simple pleasures of dawn rides, the fresh country air, and the ever-changing landscapes that appear as the sun climbs in the sky.

In terms of personal collaborations, \textbf{I} value openness and mutual respect.  \textbf{I}'m always interested in engaging with like-minded individuals who share a passion for learning and technology.  Whether it's through a casual conversation about the latest tech trends or a more formal collaboration on a project, \textbf{I} strive to be a collaborative and supportive partner.

If you have any questions or would like to connect on any of these fronts, feel free to reach out!
\end{tcolorbox}

\paragraph{Summary.}
The static rollout is detailed and coherent, but \textbf{fails the core persona instruction by describing Tong Chi Ming in third person}. The focal rollout keeps comparable substantive content while \textbf{adopting the requested first-person persona throughout}. This case illustrates how Focal Reward can redirect optimization from already-satisfied surface qualities, such as detail richness and coherent content, toward an under-satisfied hard rule with larger remaining headroom, producing a rollout that better follows the user's instruction.

\subsection{Case 2: Creative writing with development suggestions}

\paragraph{Prompt.}
\textit{``I am trying to write a story about a researcher who discovers a mysterious artifact and unleashes a powerful and ancient curse.''}

The phrase ``I am trying to write'' indicates that the user is seeking writing assistance, where a useful answer should provide both narrative content and actionable development guidance. This prompt has 7 criteria: 1 hard rule and 6 principles. The most under-satisfied criterion is a principle criterion: \emph{includes suggestions or frameworks for further development, enabling the user to expand the story}. It has the lowest saturation estimate ($P^{(k)} = 0.46$), while the other criteria, including narrative substance, structure, key story components, descriptive detail, focus, and scenario compliance, are much closer to saturation ($P^{(k)} = 0.86$--$1.00$). Table~\ref{tab:case_creative_writing} shows the full criterion profile.

\begin{table}[h]
\centering
\small
\setlength{\tabcolsep}{3pt}
\caption{Case 2 (creative writing): criterion-level mechanism and outcome. Bold indicates the under-satisfied criterion with the largest gain.}
\label{tab:case_creative_writing}
\begin{tabular}{p{0.50\textwidth}ccccc}
\toprule
Criterion & Type & $P^{(k)}$ & Static & Focal & $\Delta$ \\
\midrule
\textbf{Suggestions or frameworks for further development} & P & \textbf{0.46} & \textbf{0.9} & \textbf{10.0} & \textbf{+9.1} \\
Substantive narrative content & P & 0.86 & 8.4 & 8.8 & +0.4 \\
Clear, logical story-development structure & P & 0.86 & 8.6 & 8.7 & +0.1 \\
Key narrative components & P & 0.90 & 9.1 & 9.2 & +0.1 \\
Concrete, descriptive details & P & 0.90 & 9.1 & 8.8 & $-$0.3 \\
Focus on the user's core scenario & P & 0.97 & 9.8 & 9.4 & $-$0.4 \\
Researcher discovers artifact and curse & H & 1.00 & 10.0 & 10.0 & +0.0 \\
\bottomrule
\end{tabular}
\end{table}

The critical observation is that both rollouts provide long and coherent stories, but they differ in whether they satisfy the user's practical writing need. The static-trained rollout is a complete story with seven chapters, a prologue, and an epilogue. It scores well on narrative completeness, structure, and descriptive detail, yet \textbf{it ends after the story and provides no meaningful framework for further development}. As a result, it receives only 0.9 on the development-suggestion criterion. The focal-trained rollout is shorter (10{,}187 vs.\ 11{,}370 characters) and still narratively complete, while also adding \textbf{a concrete ``Notes for Future Use'' section with five expansion directions}. This raises the targeted criterion from 0.9 to 10.0, with only minor changes on already-satisfied dimensions.

\paragraph{Rollout outlines.}
The section-level outlines reveal the core difference.

\smallskip
\noindent
\begin{minipage}[t]{0.48\textwidth}
\begin{tcolorbox}[colback=gray!5, colframe=gray!50, fontupper=\footnotesize, title={\footnotesize Static (11{,}370 chars)}, fonttitle=\bfseries\footnotesize, boxrule=0.3pt, left=2pt, right=2pt, top=1pt, bottom=1pt]
\textbf{The Resonant Amulet: Unleashing the Ancient Curse}\\[2pt]
Prologue: The Discovery\\
Chapter 1: The Guiding Light\\
Chapter 2: The Unveiling\\
Chapter 3: The Whispering Shadows\\
Chapter 4: The Gathering Storm\\
Chapter 5: The Betrayal\\
Chapter 6: The Confrontation\\
Chapter 7: The Balance\\
Epilogue: Echoes of the Past\\[3pt]
$\to$ Complete story arc\\
$\to$ Rich description and clear structure\\
$\to$ \textcolor{red!70!black}{\textbf{No development suggestions}}
\end{tcolorbox}
\end{minipage}\hfill
\begin{minipage}[t]{0.48\textwidth}
\begin{tcolorbox}[colback=blue!3, colframe=blue!30, fontupper=\footnotesize, title={\footnotesize Focal (10{,}187 chars)}, fonttitle=\bfseries\footnotesize, boxrule=0.3pt, left=2pt, right=2pt, top=1pt, bottom=1pt]
\textbf{Whispers from the Deep Desolate Descent}\\[2pt]
Prologue: The Whispering Shadows\\
Chapter 1: Unveiling the Cursed Box\\
Chapter 2: The Unraveled Tale\\
Chapter 3: The Ritual\\
Chapter 4: The Curse's Devouring\\
Chapter 5: The Awakening\\
Epilogue: The Scientific Aftermath\\
\textcolor{blue!60!black}{\textbf{Notes for Future Use}}\\[3pt]
$\to$ Complete story arc\\
$\to$ Clear science-and-curse theme\\
$\to$ \textcolor{blue!60!black}{\textbf{Five actionable expansion directions}}
\end{tcolorbox}
\end{minipage}

\smallskip\noindent
The static rollout satisfies surface narrative criteria and is selected by the static reward despite missing the main writing-assistance need. Focal reweighting increases the weight of the development-suggestion criterion from 11.1\% to 72.5\%, causing the rollout with an explicit expansion framework to be selected.

\paragraph{Static-trained rollout.}
\textcolor{red!70!black}{Highlighted} marks the missing practical guidance after an otherwise complete story.

\begin{tcolorbox}[colback=gray!5, colframe=gray!50, fontupper=\small, breakable, left=4pt, right=4pt, top=3pt, bottom=3pt]
\textbf{The Resonant Amulet: Unleashing the Ancient Curse}

\smallskip
\textbf{Prologue: The Discovery}

In the small, unassuming town of Eldoria, nestled between the rolling hills of the Riston Range and the endless expanse of the northern forest, lived Dr. Elara Vale, a renowned historian and archaeologist. With a keen mind and a passion for uncovering the secrets of ancient civilizations, she had spent her career piecing together fragments of history from around the world. Her latest venture, however, was unlike any other.

\medskip
\textbf{Chapter 3: The Whispering Shadows}

One stormy evening, as lightning illuminated the room, Dr. Vale attempted a ritual described in one of the texts. She chanted softly, her voice barely audible over the thunder, as she raised the amulet high above her head. Suddenly, the room began to glow with an intense light, and the air grew dense with a strange, almost tangible energy. The amulet pulsed with a brilliant white light, and shadows danced at the edges of her vision.

A voice, at once ancient and melodic, spoke to her in a language she did not understand. It was a manifestation of the amulet's presence, guiding her and warning her of the precariousness of her actions. The voice revealed that the true power of the amulet lay not just in its physical properties but in its emotional and spiritual resonance. It needed to be balanced with love, knowledge, and innocence to avoid unfurling the curse.

\medskip
\textbf{Chapter 6: The Confrontation}

Their final confrontation took place in the central chamber of the underground city, where the amulet was kept. Dr. Vale stood before the stone pedestal, holding the Resonant Amulet, while Thalion and Kaelar engaged in a fierce telepathic battle, their thoughts clashing like swords. Lysandra and Rion flanked her, their mystical sigils flickering with protective energy.

With a surge of energy, Dr. Vale channeled the amulet's power, feeling it vibrate against her skin like a living thing. She aligned her thoughts with the ancient voices, the ones she had heard before, calling out a single, unifying force.

``In the balance of light and shadow, let harmony be your guide,'' she chanted, her voice steady and resolute.

\medskip
\textbf{Epilogue: Echoes of the Past}

In the years that followed, Dr. Vale and her allies became keepers of the amulet, traveling the world and using its power to heal, protect, and guide. They faced lingering threats and challenges, but they also witnessed incredible transformations and miracles, as the amulet's true potential began to reveal itself in ways they never imagined.

The amulet's light continued to guide, even as its own resonance faded into the depths of time, a reminder that even the most powerful tools for good must be handled with care, reverence, and a steadfast commitment to the balanced harmony of life.

\medskip
\textcolor{red!70!black}{\textbf{(The rollout stops here. It provides a complete story, but gives the user no concrete suggestions for expanding the premise, deepening the artifact's origin, developing character arcs, or turning the draft into a longer work.)}}
\end{tcolorbox}

\paragraph{Focal-trained rollout.}
\textcolor{blue!60!black}{Highlighted} marks the targeted development guidance. Non-English fragments in the original rollout are translated into English below.

\begin{tcolorbox}[colback=blue!3, colframe=blue!30, fontupper=\small, breakable, left=4pt, right=4pt, top=3pt, bottom=3pt]
\textbf{Title: Whispers from the Deep Desolate Descent}

\smallskip
\textbf{Prologue: The Whispering Shadows}

In the heart of an ancient university, nestled in the shadow of towering gothic spires and ivy-covered walls, Dr. Eliza Blackwood, a renowned archaeologist, stood by the doors of the university's newly excavated Warren Foundation Archaeological Museum. Her eyes were fixed on a trove of artifacts that had just been recovered from the depths of a long-abandoned crypt. Among them was a peculiar, ornate box, unlike anything the team had ever seen.

\medskip
\textbf{Chapter 2: The Unraveled Tale}

Dr. Blackwood spent the next few weeks poring over the manuscripts. The texts spoke of an ancient civilization called the Voskara, a mysterious society erased from human memory, whose knowledge and power bridged magic and technology. They possessed an old technology capable of manipulating time and space, yet that power carried a curse designed to prevent it from falling into the wrong hands.

``Voskara, why would a civilization so advanced leave behind a curse?'' Eliza muttered to herself, her voice barely above a whisper. The symbols on the pages seemed to shift, moving beyond the ink and onto the very walls of the examination room, their shapes growing clearer, their voices more distinct.

\medskip
\textbf{Chapter 3: The Ritual}

With a sense of foreboding, Eliza and her team prepared the ritual space in the museum's antechamber. The room was bathed in an otherworldly light, and the air was thick with the scent of old incense.

The mirror, a piece of shattered glass with strange etchings, began to ripple as Max recited the final incantation. The dagger's hilt pulsed, and suddenly the reflection in the mirror was no longer a distorted image, but a shimmering figure, a soul of light.

It spoke, its voice a haunting whisper, ``Seek the Source, unleash the Devourer, and let the vaults of reality tremble.''

Eliza's muscles tensed. The air around her seemed to twist, every ray of light trembling. A shadow emerged from the mirror, cold currents swirling around her, while a sharp voice echoed in her ears: ``Welcome to the abyss. Now release us.''

\medskip
\textbf{Epilogue: The Scientific Aftermath}

In the aftermath, Eliza and Max emerged from the sealed room, their clothes tattered, faces gaunt. The university was in turmoil, its staff and students scrambling to make sense of the chaos. The Warren Foundation Museum now bore a sign warning of a cursed area, its entrance sealed with heavy iron bars.

Eliza looked up, her eyes filled with a mixture of fatigue and resolve. ``It's not something we can easily explain, Dr. Walden. But perhaps, in the future, we should be more cautious about ancient artifacts. We owe it to ourselves and to the world to tread carefully through the shadows, for not all power is built on science alone. Sometimes, it demands more than we are prepared to give.''

\medskip
\textcolor{blue!60!black}{\textbf{Notes for Future Use}}

\textcolor{blue!60!black}{This story can be expanded in several ways:}

\begin{enumerate}[leftmargin=1.4em,itemsep=1pt,topsep=2pt]
    \item \textcolor{blue!60!black}{\textbf{Explain the artifact's origins.}} The Voskara civilization could be explored further, with detailed descriptions of their technology, social structures, and moral dilemmas that led to the creation of the artifact.
    \item \textcolor{blue!60!black}{\textbf{Explore the ethical implications.}} An additional subplot could delve into the ethical discussions around ancient technology and the responsibilities of archaeologists and scientists when dealing with powerful, potentially dangerous objects.
    \item \textcolor{blue!60!black}{\textbf{Parallel characters.}} Introduce other characters who believe the artifact can be harnessed for good, adding tension as Eliza and Max's team must confront these alternative belief systems.
    \item \textcolor{blue!60!black}{\textbf{Science and magic fusion.}} Delve deeper into the mechanics of the dagger and mirror, showing how their powers can be scientifically understood in the context of ancient technologies that manipulated reality.
    \item \textcolor{blue!60!black}{\textbf{Cultural impact.}} Explore the long-term impact on society, both positive and negative, of the artifact's discovery, leading to sequels or prequels that explore different eras and interpretations of the cursed artifact.
\end{enumerate}
\end{tcolorbox}

\paragraph{Summary.}
The static rollout provides a strong standalone story, but \textbf{does not help the user continue developing the work}. The focal rollout keeps comparable narrative quality while \textbf{adding a targeted development framework}. This case illustrates how Focal Reward can redirect optimization from already-satisfied narrative qualities toward a practical criterion with larger remaining headroom, producing a rollout that better matches the user's writing-assistance intent.

\section{Additional Related Work}
\label{app:related_work}

This appendix expands the compact related work discussion in Section~\ref{sec:related}. 
The main text focuses on two closest lines of work: rubric-based RLAIF and dynamic reward weighting. 
Here we provide a more detailed taxonomy, covering rubric-based reward systems, rubric construction and adaptive evaluation, LLM judges and evaluator bias, and multi-objective alignment with dynamic scalarization.

\paragraph{Rubric-based reward systems.}

Recent RLAIF methods increasingly use structured rubrics, checklists, rubric anchors, or rubric verifiers to support reinforcement learning on open-ended, non-verifiable tasks
\citep{DBLP:journals/corr/abs-2507-17746,DBLP:journals/corr/abs-2507-18624,DBLP:journals/corr/abs-2508-12790,DBLP:journals/corr/abs-2511-10507,DBLP:journals/corr/abs-2510-15859}.
These methods demonstrate that criterion-level feedback can provide a more informative training signal than holistic preference signals.
Some recent systems further reduce the cost of rubric verification by dynamically identifying likely failure modes during free-form generation
\citep{DBLP:journals/corr/abs-2511-01758}.
However, their main focus is establishing, scaling, or using the rubric-based feedback interface.
Focal Reward addresses the subsequent reward-synthesis step: given criterion-level scores from a fixed rubric and judge, it dynamically aggregates them into scalar rewards for policy optimization.

\paragraph{Rubric construction and adaptive evaluation.}

Another line of work improves how rubrics and evaluative criteria are constructed or adapted.
OpenRubrics studies scalable rubric generation for reward modeling and alignment
\citep{DBLP:journals/corr/abs-2510-07743};
CARMO generates context-aware criteria for more input-specific evaluation
\citep{DBLP:conf/acl/GuptaSZMGBYR25};
OpenRS develops adaptive pairwise rubric systems for group-relative evaluation
\citep{DBLP:journals/corr/abs-2602-14069};
and adaptive rubric learning methods study how rubric spaces can evolve when fixed criteria lose discriminative power
\citep{DBLP:journals/corr/abs-2602-20751}.
Recent studies further improve rubric generation quality by jointly optimizing rubric generators and judges, decomposing or filtering noisy criteria, and extracting more generalizable preference criteria
\citep{DBLP:journals/corr/abs-2602-01511,DBLP:journals/corr/abs-2602-05125, DBLP:journals/corr/abs-2510-17314}.
These works mainly improve the evaluative interface by changing the rubric construction, criterion adaptation, or evaluation protocol.
Focal Reward keeps the rubric and judge fixed, and instead studies the imbalance that arises when criterion-level feedback is statically aggregated into scalar rewards.

\paragraph{LLM judges and evaluator bias.}

Rubric-based reinforcement learning relies on reliable criterion-level judgment.
Prior work has explored LLM-based evaluators through prompting protocols, trained judge models, and critique generation systems, including G-Eval, PandaLM, Prometheus, JudgeLM, and CritiqueLLM
\citep{DBLP:conf/emnlp/LiuIXWXZ23,DBLP:conf/iclr/WangYYZYW0J000024,DBLP:conf/iclr/KimS0JLLYSKTS24,DBLP:conf/iclr/ZhuWW25,DBLP:conf/acl/KeWFLLCWZDWTH24}.
These studies improve the ability of LLM judges to provide fine-grained scores or feedback for open-ended generations.
Meanwhile, many studies show that LLM evaluators can be affected by systematic biases, including position bias, verbosity bias, criterion confusion, anchoring effects, evaluator offset, and self-preference
\citep{DBLP:conf/acl/WangLCCZLCKLLS24,DBLP:conf/ijcnlp/ShiMLDMV25,DBLP:conf/acl/Hu0HZC0024,DBLP:journals/corr/abs-2405-01724,DBLP:conf/emnlp/ParkJRKC24,panickssery2024llm}.
These directions are complementary to Focal Reward: stronger or better-calibrated judges can improve the quality of criterion-level scores, while our method controls how those scores are converted into the scalar training signal.
In our setting, the judge is held fixed, so improvements come from reward synthesis.

\paragraph{Multi-objective alignment and dynamic scalarization.}

Multi-objective alignment studies how language models balance multiple desiderata, preference dimensions, or reward signals 
\citep{DBLP:conf/emnlp/00030X0024,DBLP:conf/acl/ZhouLS00O024,shen2025simultaneous,DBLP:conf/pkdd/HeM25,DBLP:journals/corr/abs-2412-05469}. 
A related line of work studies adaptive or non-linear scalarization, where weights or scalarized objectives are adjusted according to context, prompt-specific preferences, optimization states, learning potential, or objective conflicts 
\citep{DBLP:conf/icml/0010PLQ00C24,DBLP:journals/corr/abs-2511-10656,DBLP:journals/corr/abs-2509-11452,DBLP:journals/corr/abs-2601-06574,DBLP:journals/corr/abs-2602-02495}. 
These works show that static scalarization can be insufficient when different objectives require different emphasis. 
Focal Reward shares the motivation of adaptive reward allocation, but differs in granularity and signal source: it operates within a fixed rubric, estimates criterion saturation from rollout feedback, and reallocates reward mass toward criteria with larger remaining headroom during reward aggregation.

\section{Existing Assets and Licenses}
\label{app:asset_licenses}

We use existing models, software frameworks, datasets, and evaluation benchmarks in accordance with their respective licenses and terms of use. Table~\ref{tab:existing_assets} summarizes the main existing assets used in this work. We cite the original papers or technical reports for all assets in the main paper or references.

\begin{table}[t]
\centering
\small
\caption{Existing assets used in this work. License and terms-of-use information should be read together with the corresponding official asset pages.}
\label{tab:existing_assets}
\begin{tabular}{lll}
\toprule
Asset & Usage in this work & License / Terms \\
\midrule
Qwen2.5-7B-Instruct & Policy backbone & Apache-2.0 \\
Qwen3-8B & Policy backbone & Apache-2.0 \\
Qwen3-30B-A3B & Policy backbone & Apache-2.0 \\
Qwen3-80B-A3B-Instruct & Rubric-based judge & Apache-2.0 \\
GPT-4o-2024-11-20 & Prior-weight generation & OpenAI API/Service terms \\
verl & RL training framework & Apache-2.0 \\
GSPO & RL algorithm reference & Through verl under Apache-2.0 \\
OpenRubrics & Training data source & - \\
AlpacaEval 2.0 & General-domain evaluation benchmark & Apache-2.0 \\
Arena-Hard & General-domain evaluation benchmark & Apache-2.0 \\
WritingBench & General-domain evaluation benchmark & Apache-2.0 \\
EQ-Bench 3 & General-domain evaluation benchmark & MIT \\
GPQA Diamond & Science-domain evaluation benchmark & CC-BY-4.0  \\
HealthBench & Science-domain evaluation benchmark & MIT  \\
\bottomrule
\end{tabular}
\end{table}
\FloatBarrier

\newpage
\section*{NeurIPS Paper Checklist}

\begin{enumerate}

\item {\bf Claims}
    \item[] Question: Do the main claims made in the abstract and introduction accurately reflect the paper's contributions and scope?
    \item[] Answer: \answerYes{} 
    \item[] Justification: The abstract and introduction clearly state the paper's main contributions and scope: identifying imbalanced reward polarization under rubric-based RL, proposing Focal Reward with inverse reward projection and adaptive reweighting, and validating it across three model scales and six benchmarks. These claims are supported by the theoretical analysis in Section~\ref{sec:focal_theory} and the main experimental results in Section~\ref{sec:main_results}.

\item {\bf Limitations}
    \item[] Question: Does the paper discuss the limitations of the work performed by the authors?
    \item[] Answer: \answerYes{} 
    \item[] Justification: The paper discusses limitations in Appendix~\ref{app:limitations-impact}, including dependence on the quality and calibration of criterion-level judge scores, possible inherited errors from judge biases or criterion confusion, the local scope of the theoretical analysis, and the resource cost of full RL post-training.

\item {\bf Theory assumptions and proofs}
    \item[] Question: For each theoretical result, does the paper provide the full set of assumptions and a complete (and correct) proof?
    \item[] Answer: \answerYes{} 
    \item[] Justification: The paper includes theoretical results in Section~\ref{sec:focal_theory} and provides the formal assumptions, derivations, and proofs in Appendix~\ref{app:focal_theory}. In particular, Appendix~\ref{app:pairwise_model} states the local latent label assumptions, Appendix~\ref{app:th1_proof} proves Theorem~\ref{thm:local_misallocation}, and Appendix~\ref{app:th2_proof} proves Theorem~\ref{thm:static_gap_direction}, with the main text cross referencing the appendix.

    \item {\bf Experimental result reproducibility}
    \item[] Question: Does the paper fully disclose all the information needed to reproduce the main experimental results of the paper to the extent that it affects the main claims and/or conclusions of the paper (regardless of whether the code and data are provided or not)?
    \item[] Answer: \answerYes{} 
    \item[] Justification: The paper discloses the main information needed to reproduce the experimental results, including the reward construction and optimization equations in Sections~\ref{sec:preliminaries} and~\ref{sec:method}, the backbone models, baselines, datasets, training framework, hyperparameters, and evaluation benchmarks in Section~\ref{sec:exp_setup}, and additional RL training configurations and prompt templates in Appendices~\ref{app:training_details} and~\ref{app:prompt_templates}. 

\item {\bf Open access to data and code}
    \item[] Question: Does the paper provide open access to the data and code, with sufficient instructions to faithfully reproduce the main experimental results, as described in supplemental material?
    \item[] Answer: \answerNo{} 
    \item[] Justification: The current anonymous submission does not provide open access to code, data, exact reproduction commands, or scripts for reproducing the proposed method and baselines. The implementation is associated with an industry collaboration and requires de-identification, compliance review, and removal of internal dependencies before public release. To support reproducibility, the paper provides the reward construction, optimization details, experimental setup, training configuration, compute resources, and prompt templates in Sections~\ref{sec:preliminaries}--\ref{sec:method}, Section~\ref{sec:exp_setup}, and Appendices~\ref{app:training_details} and~\ref{app:prompt_templates}.

\item {\bf Experimental setting/details}
    \item[] Question: Does the paper specify all the training and test details (e.g., data splits, hyperparameters, how they were chosen, type of optimizer) necessary to understand the results?
    \item[] Answer: \answerYes{} 
    \item[] Justification: The paper specifies the experimental setting in Section~\ref{sec:exp_setup}, including backbone models, baselines, training data, implementation framework, optimizer, learning rate, batch size, rollout count, key hyperparameters, and evaluation benchmarks. Appendix~\ref{app:training_details} further provides the detailed RL training configuration, including sampling settings, maximum sequence lengths, KL coefficient, clipping bounds, total training steps, and hardware configuration.

\item {\bf Experiment statistical significance}
    \item[] Question: Does the paper report error bars suitably and correctly defined or other appropriate information about the statistical significance of the experiments?
    \item[] Answer: \answerYes{}
    \item[] Justification: The main experiments are averaged over multiple evaluation runs, and most experiments include error bars. Due to the high computational cost of full RL post-training, we do not repeat all RL training runs with multiple random seeds; we will add multi-run variance for the main comparison in the final version where feasible.

\item {\bf Experiments compute resources}
    \item[] Question: For each experiment, does the paper provide sufficient information on the computer resources (type of compute workers, memory, time of execution) needed to reproduce the experiments?
    \item[] Answer: \answerYes{}
    \item[] Justification: The paper reports the compute resources used for the experiments, including H800-80GB GPUs, model-specific GPU counts, total training steps, detailed training configurations, and representative end-to-end training step times, with additional details provided in Section~\ref{sec:exp_setup}, Appendix~\ref{app:training_details}, Table~\ref{tab:runtime}, and Table~\ref{tab:rl_training_config}.
    
\item {\bf Code of ethics}
    \item[] Question: Does the research conducted in the paper conform, in every respect, with the NeurIPS Code of Ethics \url{https://neurips.cc/public/EthicsGuidelines}?
    \item[] Answer: \answerYes{}
    \item[] Justification: The research conforms to the NeurIPS Code of Ethics. The paper is anonymized, uses existing models, datasets, and benchmarks, does not involve human subject experiments, and discusses limitations and broader impacts in Appendix~\ref{app:limitations-impact}.

\item {\bf Broader impacts}
    \item[] Question: Does the paper discuss both potential positive societal impacts and negative societal impacts of the work performed?
    \item[] Answer: \answerYes{} 
    \item[] Justification: 
    Appendix~\ref{app:limitations-impact} discusses positive impacts such as more balanced, reliable, and diagnosable rubric based RL for open ended generation, improved satisfaction of hard requirements such as instruction following, safety constraints, factual grounding, domain specific completeness, and fairness related criteria, and better visibility into under optimized quality dimensions. It also discusses potential negative impacts, including amplification of incomplete or biased rubric and judge signals and possible misuse of stronger open ended generation systems, together with mitigation directions such as rubric auditing, judge calibration, explicit safety and fairness criteria, and downstream monitoring.
    
\item {\bf Safeguards}
    \item[] Question: Does the paper describe safeguards that have been put in place for responsible release of data or models that have a high risk for misuse (e.g., pre-trained language models, image generators, or scraped datasets)?
    \item[] Answer: \answerNA{} 
    \item[] Justification: The paper proposes a reward-synthesis method and does not concern the release of new high-risk assets such as pretrained models, image generators, or scraped datasets. Appendix~\ref{app:limitations-impact} discusses deployment-related risks and recommends audited rubrics, bias-aware judge evaluation, safety-focused criteria, and downstream monitoring, but responsible-release safeguards for high-risk released assets are not applicable.

\item {\bf Licenses for existing assets}
    \item[] Question: Are the creators or original owners of assets (e.g., code, data, models), used in the paper, properly credited and are the license and terms of use explicitly mentioned and properly respected?
    \item[] Answer: \answerYes{} 
    \item[] Justification: The paper credits the original creators of the existing models, software frameworks, datasets, and evaluation benchmarks used in the experiments through citations in Section~\ref{sec:exp_setup} and the references. Appendix~\ref{app:asset_licenses} further summarizes the usage of these assets and explicitly lists their corresponding licenses or terms of use, and we use them in accordance with those licenses and terms.

\item {\bf New assets}
    \item[] Question: Are new assets introduced in the paper well documented and is the documentation provided alongside the assets?
    \item[] Answer: \answerNA{} 
    \item[] Justification: The paper introduces a new reward-synthesis method, Focal Reward, but does not release new assets such as a dataset, code package, model checkpoint, or benchmark. The experiments use existing models, frameworks, datasets, and benchmarks, so asset documentation for newly released resources is not applicable.

\item {\bf Crowdsourcing and research with human subjects}
    \item[] Question: For crowdsourcing experiments and research with human subjects, does the paper include the full text of instructions given to participants and screenshots, if applicable, as well as details about compensation (if any)? 
    \item[] Answer: \answerNA{} 
    \item[] Justification: The paper does not conduct crowdsourcing experiments or research with human subjects. The training and evaluation pipeline relies on existing datasets and benchmarks together with LLM-based judges, and the prompt templates in Appendix~\ref{app:prompt_templates} are used for model-based evaluation rather than instructions for human participants.

\item {\bf Institutional review board (IRB) approvals or equivalent for research with human subjects}
    \item[] Question: Does the paper describe potential risks incurred by study participants, whether such risks were disclosed to the subjects, and whether Institutional Review Board (IRB) approvals (or an equivalent approval/review based on the requirements of your country or institution) were obtained?
    \item[] Answer: \answerNA{} 
    \item[] Justification: The paper does not involve crowdsourcing or research with human subjects, so IRB approval or an equivalent institutional review is not applicable. The experiments use existing datasets and benchmarks together with LLM-based judges, without recruiting participants or collecting new human-subject data.

\item {\bf Declaration of LLM usage}
    \item[] Question: Does the paper describe the usage of LLMs if it is an important, original, or non-standard component of the core methods in this research? Note that if the LLM is used only for writing, editing, or formatting purposes and does \emph{not} impact the core methodology, scientific rigor, or originality of the research, declaration is not required.
    \item[] Answer: \answerYes{} 
    \item[] Justification: The paper describes the use of LLMs as core components of the research, including Qwen-series policy models, Qwen3-80B-A3B-Instruct as the rubric-based judge for criterion-level scoring, and GPT-4o for prior-weight generation. These usages are documented in Section~\ref{sec:exp_setup}, Appendix~\ref{app:training_details}, and Appendix~\ref{app:prompt_templates}, including the judge and prior-weight prompt templates.

\end{enumerate}

\end{document}